\documentclass[11pt]{article}

\usepackage[final]{acl}
\usepackage{xcolor}
\usepackage{float}
\usepackage{comment}
\usepackage{microtype}
\usepackage{amsfonts}
\usepackage{amsmath} 
\usepackage{booktabs}
\usepackage{subcaption}
\usepackage{graphicx}
\usepackage{times}
\usepackage{latexsym}
\usepackage{booktabs}
\usepackage{multirow}
\usepackage[T1]{fontenc}

\usepackage[utf8]{inputenc}

\usepackage{microtype}

\usepackage{inconsolata}

\usepackage{graphicx}

\definecolor{mygreen}{RGB}{86, 188, 108}

\title{One Mask to Rule Them All: \\ On Hidden Facts after Editing and How to Find Them}

\author{
   Ali Holmov $^{1, 2}$ \quad
   Paul Youssef $^2$ \quad
   Nandi Schoots $^3$ \quad
   Christin Seifert $^2$
\\ 
\textsuperscript{1} Technical University of Munich,
\textsuperscript{2} Marburg University,
\textsuperscript{3} University of Oxford
\\
\texttt{ali.kholmovaia@tum.de}, \texttt{nandischoots@gmail.com} \\
\texttt{\{paul.youssef, christin.seifert\}@uni-marburg.de}
}

\begin{document}
\maketitle
\begin{abstract}
Knowledge editing methods such as ROME and MEMIT update factual associations in transformer models by modifying MLP weights. While evaluated mainly by output behavior, their internal mechanism remains underexplored. We investigate whether edits rely on a common mechanism, regardless of which fact is modified. Despite fact-specific weight changes, we argue that ROME and MEMIT target the same subset of weights that are critical for maintaining edits. To isolate this subset, we train a compact binary mask over the edited weights. The mask reverses 80\% of edits on the training set and over 70\% on the test set, confirming that diverse edits share a common functional structure. Our analysis reveals that the mask reverses edits by eliminating overattention in later layers. Additionally, we show that injecting the mask during editing drops editing success from 98\% to 38\%, demonstrating that this mechanism is necessary for edits to succeed. Our finding that edits suppress rather than overwrite knowledge explains why ROME and MEMIT fail to propagate changes to related facts. The identified common functional subspace informs detection and defense against unwanted edits\footnote{Code available at: \href{https://github.com/holmov1/one-mask-ke}{github.com/holmov1/one-mask-ke}}.
\end{abstract}

\section{Introduction}
\begin{figure}[t]
    \centering
    \includegraphics[width=1.0\columnwidth, clip, trim={1.9cm 0.1cm 0.7cm 0.6cm}]{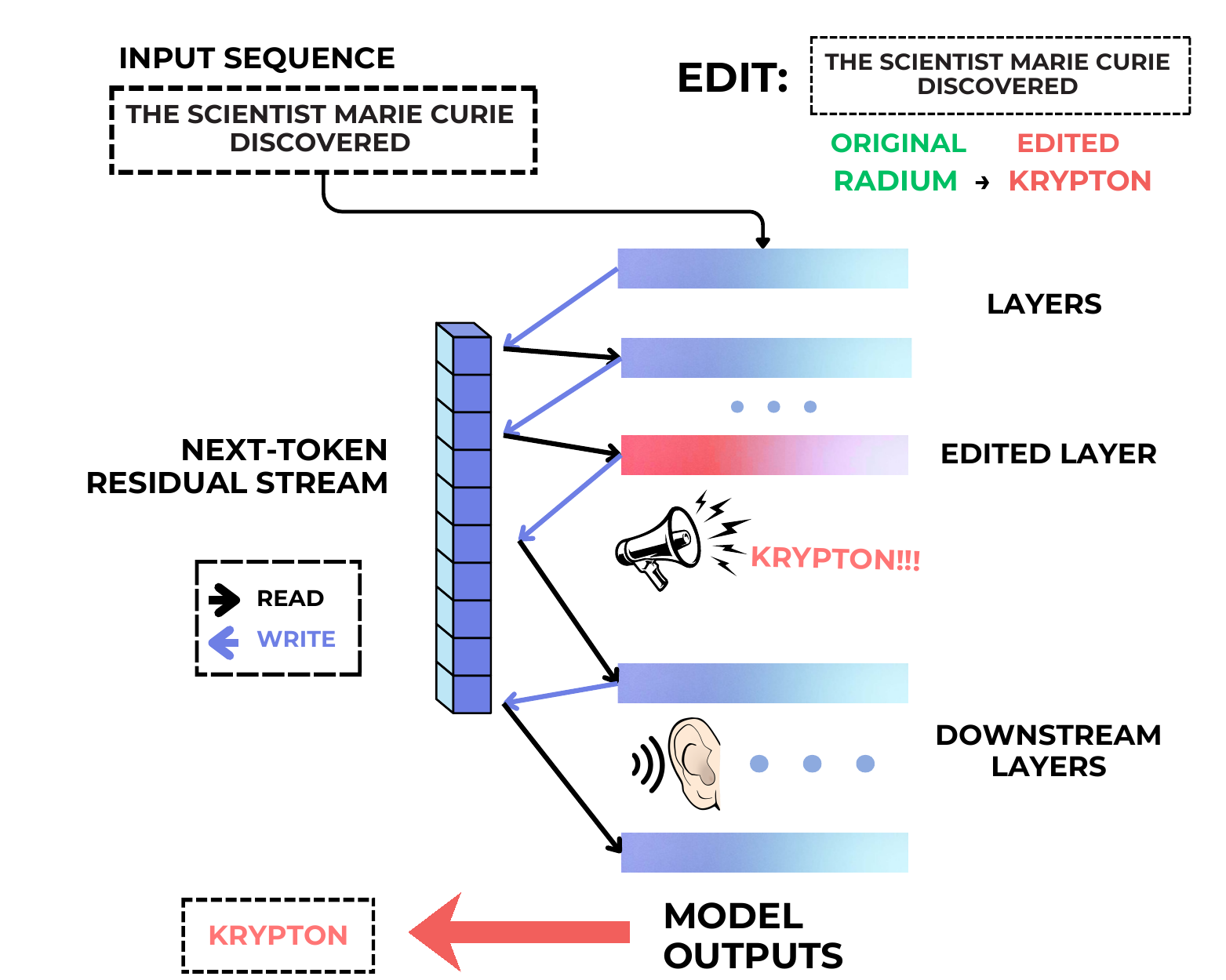}
\caption{\textbf{Edits succeed by hijacking attention, not by overwriting knowledge.}
When a locate-and-edit method (ROME/MEMIT) replaces an original object (\textcolor{mygreen}{\textsc{radium}}) with an edited one (\textcolor{red}{\textsc{krypton}}), the modified MLP weights at the edited layer write an amplified signal into the residual stream. Downstream layers attend disproportionately to this signal, suppressing the retrieval of the original fact rather than erasing it.}
\label{fig:hypothesis}
\end{figure}

Knowledge Editing (KE)~\cite{wang-etal-2024:ACMSurvey} aims to update specific facts in transformer models without expensive retraining. Among KE methods, locate-and-edit approaches such as ROME~\cite{meng2022locating} and MEMIT~\cite{meng2023massediting} have gained traction for their efficiency. These methods identify parameters associated with target facts and directly modify MLP weights to overwrite factual associations. These methods are motivated by the hypothesis that MLP layers act as associative memories \cite{geva-etal-2021-transformer}, and that targeted updates can overwrite specific facts.

Editing success is evaluated solely based on whether the model outputs the new fact. Thus, the claim that knowledge is overwritten rests on this behavioral change alone. It remains unexplored, however, how edits affect the model's internal representations. 
Additionally, there is a conceptual puzzle: While transformers are said to retrieve knowledge through redundant pathways ~\citep{mcgrath2023hydraeffectemergentselfrepair, hase2023does}, KE methods successfully update facts by modifying a single layer or a small range of connected layers. If factual knowledge is distributed, how can modifying a single layer or a small range of connected layers successfully override knowledge retrieval? 

We hypothesize that ROME and MEMIT succeed not by overwriting original factual knowledge, but by suppressing it, as illustrated in \autoref{fig:hypothesis}. We propose that beyond introducing fact-specific changes, these methods rely on a common subset of weights that are critical for maintaining any edit. By modifying these weights, ROME and MEMIT inject amplified signals that force the model to output edited facts and suppress the propagation of original knowledge without erasing it. These amplified signals cause the later layers to attend disproportionately to the edited signal. 

To verify this hypothesis, we isolate the subset of weights that are critical for maintaining edits by training a compact binary mask over the edited weight matrices. The mask identifies which weight changes are necessary for the edit to persist, allowing us to neutralize edits by pruning parts of the edited weights. If a compact mask suffices to remove the edit, this confirms that only a small subset of weights is critical for maintaining the edit. If the same mask removes diverse edits across different facts, this reveals a shared structure that all edits exploit. Conversely, if edits introduce fact-specific changes, no single mask should generalize across semantically different edits.

We find that a single mask that removes a small fraction of the edited 
weights (typically under 10\%) reverses over 80\% of edits on the training set and over 70\% on unseen edits. This generalization confirms that diverse edits rely on a common functional structure. Analyzing what the mask targets, we find that it eliminates amplified attention signals in later layers while preserving MLP pathways, which continue to encode the original knowledge. This reveals that edits succeed by hijacking attention rather than overwriting stored facts. Furthermore, injecting the mask prior to the editing process drops success from 98\% to 38\%, confirming that this mechanism is not just sufficient for reversal but necessary for edits to succeed. These findings suggest that ROME and MEMIT are fundamentally limited, as they cannot truly overwrite knowledge, but only suppress its retrieval. This explains their known failure to propagate changes to related facts ~\citep{hsueh-etal-2024-editing} and creates pathways for detecting and defending against unwanted edits.
\section{Related Work}
\paragraph{Editing Methods.} 
Factual knowledge in transformer language models~\cite{petroni-etal-2019-language, youssef-etal-2023-give} can become outdated. KE methods~\citep{wang-etal-2024:ACMSurvey, mazzia2024surveyknowledgeeditingneural} aim to keep facts up-to-date without expensive pre-training. KE methods fall into two categories: parameter-modifying and parameter-preserving. Parameter-modifying methods include locate-and-edit approaches such as ROME~\citep{meng2022locating} and MEMIT~\citep{meng2023massediting}, which locate and update parameters responsible for the facts, and meta-learning approaches such as MEND~\citep{mitchell2022fast} and MALMEN~\citep{tan2023massive}, which train hypernetworks to predict necessary parameter shifts to update facts. Parameter-preserving methods add memory modules~\citep{mitchell2022memory, hartvigsen2023aging, wang2024wise, guo2025balancedit} or make use of the LLMs' strong in-context abilities~\citep{cohen-etal-2024-evaluating, youssef-etal-2024-queen} to edit knowledge~\citep{zheng-etal-2023-edit, youssef-etal-2026-persuasion}. In this work, we focus on locate-and-edit KEs such as ROME and MEMIT due to their wide usage, and to better understand how these KEs adapt LLMs. 

\paragraph{Countermeasures to malicious editing.}
Research on countermeasures to editing has emerged in response to the potential malicious uses of KE. This line of work spans distinguishing edited from unedited facts via internal representations~\citep{youssef-etal-2025-fact, li2024knowledgeeditinglargelanguage}, reversing in-context and parameter-modifying edits~\citep{youssef-etal-2025-make, youssef2025tracingreversingrankonemodel}, tracing edits by training the model to decode its edited weights~\citep{youssef2025tracingreversingrankonemodel} and studying the effect of finetuning on edits~\citep{cheng2025finetuningeraseeditsfragile}. We extend this line of work by identifying a minimal set of common weights that maintain edits across diverse facts.

\paragraph{Knowledge Representation in Transformers.}
Transformers exhibit emergent self-repair capabilities: when a layer is ablated, downstream layers compensate by increasing their contributions, partially restoring original outputs \cite{mcgrath2023hydraeffectemergentselfrepair}. This redundancy implies that factual knowledge can be retrieved through multiple pathways. Work on knowledge-critical subnetworks \cite{bayazit-etal-2024-discovering} demonstrates that sparse subnetworks spanning multiple layers are responsible for maintaining specific factual associations. These findings create a paradox for locate-and-edit methods: if knowledge is distributed across redundant pathways, how can modifying a single layer or a small range of consecutive layers successfully override factual retrieval? We show that ROME and MEMIT edit facts by inducing overattention --- a common mechanism that suppresses redundant pathways without erasing them.

\paragraph{Attention Phenomena in Knowledge Editing.}
Recent work documents attention-related phenomena in edited models. \citet{wang2025revealing} identify \textit{attention drift}: excessive attention scores assigned to edited entities causing specificity failure, where edits corrupt unrelated knowledge. They propose to selectively constrain drifting attention heads via regularization during editing. \citet{xie-etal-2025-revealing-deceptiveness} analyze \textit{superficial editing}: the tendency of edited models to revert to original knowledge under adversarial prompts. They identify two contributing factors: the residual stream at the last subject position in earlier layers, and specific attention heads in later layers. Both works document how attention mechanisms relate to editing failures. Neither addresses what makes edits \textit{succeed} in the first place: the structural mechanism that suppresses original knowledge during normal retrieval. Our work identifies this mechanism and shows that it is shared across edits.
\section{Methodology}
\label{sec:method}
\begin{figure*}[h]
    \includegraphics[width=1.0\textwidth, clip, trim={0.9cm 0cm 0.85cm 0.0cm}]{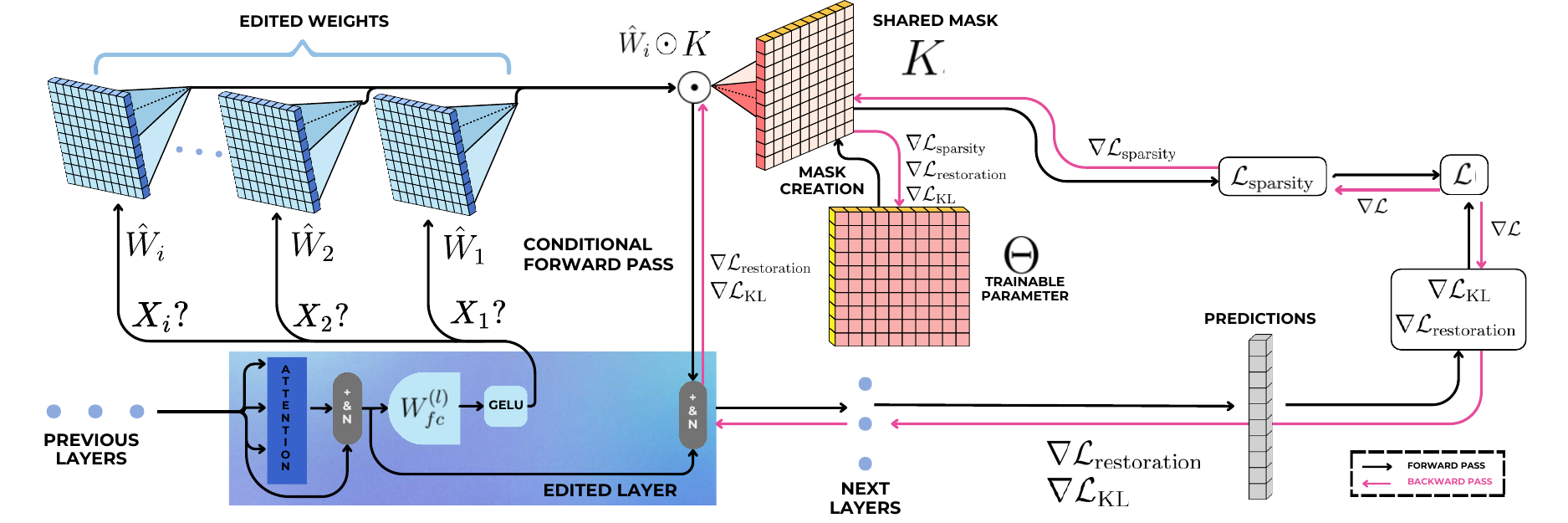}
    \caption{
        \textbf{Shared-mask training.} A single mask $K$, parameterized by trainable parameters $\Theta$, is learned jointly across $N$ edits with frozen edited weight matrices $\{\hat{W}_1, \dots, \hat{W}_N\}$. \textbf{Forward pass (black):} For each training sample $X_i$, the corresponding edited matrix $\hat{W}_i$ is routed in and combined with the shared mask via elementwise product $\hat{W}_i \odot K$. All other components (attention, remaining layers) are unchanged. \textbf{Backward pass (pink):} gradients from the combined loss $\mathcal{L}$ (Eq.~\ref{eq:loss}), such as restoration, sparsity, and KL terms, flow only to $\Theta$; edited weights and the base model stay frozen. This forces $K$ to capture structure shared across edits rather than memorize any single one.}
\label{fig:shared_mask}
  \label{fig:shared_training}
\end{figure*}
\paragraph{Background.} 
Let $M$ be the original model. A factual prompt $x$ consists of a subject $s$ and relation $r$ (e.g., ``Marie Curie discovered''), with a corresponding object $o$ (e.g., ``radium''). A knowledge edit replaces the original object $o$ with a new object $o^*$ (e.g., ``krypton''), producing an edited model $M_e$. 

\paragraph{Hypothesis.} We hypothesize that KE methods rely on a common mechanism crucial to inject and maintain edits. Specifically, beyond introducing fact-specific updates, these methods target the same subset of weights critical for the edit to persist. 

\paragraph{Mask Training.} To verify this hypothesis, we identify whether such a shared subset exists by training a binary mask $K = (k_{a,b})$, where $ k_{a,b} \in \{0,1\} \; \text{for all } a,b$,  over a set of edited weights $\hat W$ to remove the edits. If only a small subset of weights is critical for maintaining an edit, then selectively removing those weights should be sufficient to restore the original model's behavior. The mask is applied element-wise to the edited weight matrix $\hat{W}$, producing a pruned model $M_p$ with weights $\hat{W} \odot K$, where $\odot$ denotes the element-wise product operation. A mask value of 1 retains the edited weight; a value of 0 removes its contribution. If the mask successfully reverses an edit, the pruned model $M_p$ assigns a higher probability to the original object $o$ than the edited object $o^*$.

We train a single mask $K$ across diverse edits spanning semantically different facts (cf. \autoref{fig:shared_training}). If this single mask generalizes, reversing not only training edits but also unseen edits, this would constitute evidence for a shared mechanism, i.e., the same weight positions would be responsible for the edit across different facts. Conversely, if each edit introduces fact-specific changes, a single mask would not generalize well, i.e., different edits would modify different weight positions, and a mask trained on one set of facts would fail to reverse edits on new (unseen) facts.

\paragraph{Loss Function.} We train the mask to satisfy three constraints. First, \emph{restoration}: the pruned model $M_p$ should prefer the original object $o$ over the edited object $o^*$. This means that after applying the mask to the edited weights, the probability $P_{M_p}(o \mid x)$ should exceed $P_{M_p}(o^* \mid x)$. If the mask successfully restores the probabilities across diverse edits, it demonstrates that the original knowledge was never erased. Second, \emph{minimality}: the mask should remove as few weights as possible. We want most mask values to be $1$, pruning only the small subset of weights responsible for maintaining the edit. This constraint ensures that we identify the specific weights that enable the edit, rather than broadly disrupting the layer. Third, \emph{behavior preservation}: the behavior of the pruned model should remain close to the original model. This ensures that the mask does not introduce arbitrary changes that happen to flip the prediction or damage the model's general language capabilities.

We formalize each constraint as a loss term. The \emph{restoration} loss measures whether the pruned model assigns a higher probability to the original object than the edited one:
\begin{equation}
\small
\label{eq:restoration-loss}
\mathcal{L}_{\text{restoration}}= -[\log P_{M_{p}} (o \mid x) - \log P_{M_{p}} (o^* \mid x)]
\end{equation}
This loss is negative when $P_{M_p}(o \mid x) > P_{M_p}(o^* \mid x)$, i.e., when the original fact is restored. We require that $\mathcal{L}_{\text{restoration}} \leq -\delta$, where the margin $\delta$ encourages confident restoration rather than marginal preference.

The sparsity loss measures the fraction of weights pruned:
\begin{equation}
\small
\mathcal{L}_{\text{sparsity}} = \frac{1}{|K|} \sum_{a=1}^m \sum_{b=1}^n  1 - k_{a,b}
\label{eq:sparsity-loss}
\end{equation}
where $k_{a,b} \in \{0,1\}$ is the mask value at position $(a,b)$. The constraint $\mathcal{L}_{\text{sparsity}} \le S_{\text{max}}$ limits pruning to at most $S_{\text{max}}$ of the layer's weights. The \emph{behavior preservation} loss measures divergence between the output distributions of the original model $M$ and the pruned model $M_p$ using KL divergence. We combine these terms into a constrained optimization problem, minimizing the KL divergence subject to constraints on restoration and sparsity:
\begin{equation}
\small
\begin{aligned}
\min_{\theta} \beta \ \mathcal{L}_{\text{KL}}^{T}  \text{ s.t. } \mathcal{L}_{\text{sparsity}} \le S_{\text{max}}\\ \text{ and } \mathcal{L}_{\text{restoration}} \le 0-\delta
\end{aligned}
\label{eq:objective}
\end{equation}

We convert these constraints into penalty terms, yielding the combined loss:
\begin{equation}
\small
\begin{aligned}
\mathcal{L}(\theta) = \beta \mathcal{L}_{\text{KL}}^{T} + \max(0, \mathcal{L}_{\text{sparsity}} - S_{\text{max}}) \\+ \max(0, \mathcal{L}_{\text{restoration}}+\delta)
\end{aligned}
\label{eq:loss}
\end{equation}
The penalty terms become activated only when the specified constraints are violated.

\section{Experiments}
\paragraph{Training Details.}
Since a binary mask itself is non-differentiable, we initialize trainable parameters $\Theta$ and apply a sigmoid function to obtain soft mask $K \in (0, 1)$ \cite{louizos2018learning, maddison2017the}. At inference, we binarize the mask using a threshold $\gamma \in (0, 1)$. 

As illustrated in \autoref{fig:shared_training}, we train the mask across multiple edits simultaneously. Each training sample corresponds to a different edit with its own frozen edited weight matrix $\hat{W}_{i}$. During the forward pass, we swap in the appropriate $\hat{W}_{i}$ for each sample and apply the shared mask elementwise: $\hat{W}_{i} \odot K$. Gradients flow only through the mask parameters $\Theta$, while all edited weights remain fixed. This conditional forward pass ensures the mask learns patterns shared across edits rather than memorizing a single edit's structure.

\paragraph{Setup.} We use the CounterFact dataset introduced as part of the ROME study~\cite{meng2022locating}. CounterFact contains counterfactual triples. We train on 3,000 samples stratified across 10 relations and test on 1,700 held-out samples from the same relations. We use EasyEdit~\cite{wang-etal-2024-easyedit} to edit models.

We evaluate on three models: GPT-2 XL (1.5B), LLaMA-3.2 (3B) \cite{llama3}, and Qwen2.5 (7B) \cite{qwen2025qwen25technicalreport}. For ROME, we train the mask across single edits. For MEMIT, which supports batch editing, we edit 1,000 facts simultaneously and train the mask on a single edited layer. For testing, we apply 1,000 different edits from the test set.

In our experiments, we aim to answer three questions: (a) Can a single mask reverse diverse edits? (b) Does the mask generalize to unseen edits from different facts? (c) Does applying the mask preserve general language modeling capabilities?

\subsection{Evaluation Metrics}
\label{subsec:evaluation}
\paragraph{Reversal Success Rate (RSR).}
To assess whether original knowledge persists after editing, we measure if the pruned model prefers the original object over the edited one. We define $\Delta r_i := P_{M_p}(o_i \mid x_i) - P_{M_p}(o^*_i \mid x_i)$ and compute RSR as the proportion of samples where $\Delta r_i > 0$. High RSR indicates that original facts survive the edit and can be recovered.

\paragraph{Top-1 Overlap (Top-1).}
While RSR measures relative preference, it does not guarantee exact behavioral restoration. We report Top-1 Overlap: the percentage of samples where the pruned model's top prediction matches the original model's. This verifies whether the model's generation path has been successfully reverted.

\paragraph{Perplexity (PPL).} Perplexity is one of the most widely used metrics for evaluating language models and is commonly applied to the WikiText datasets \cite{Radford2019LanguageMA, Huang2024}. We measure perplexity of $M$, $M_e$, and $M_p$ on a random 70k-token subset of WikiText-2 \cite{merity2017pointer} to assess whether reversals affect general language modeling capabilities. 

\begin{table}
\centering
\small
\begin{tabular}{llcc}
\toprule
\textbf{Method} & \textbf{Model} & \textbf{RSR} $\uparrow$ & \textbf{Top-1 Overlap} $\uparrow$ \\
\midrule
\multirow{3}{*}{ROME}  & GPT-2 XL      & 0.2\% & 1.0\% \\
                       & LLaMA-3.2  & 0.0\% & 0.5\% \\
                       & Qwen-2.5   & 1.7\% & 3.1\% \\
\midrule
\multirow{3}{*}{MEMIT} & GPT-2 XL      & 7.2\% & 14.4\% \\
                       & LLaMA-3.2  & 0.2\% & 0.6\% \\
                       & Qwen-2.5   & 0.0\% & 0.3\% \\
\bottomrule
\end{tabular}
\caption{RSR and Top-1 Overlap of edited models.  
 RSR measures how often the edited model still prefers the original fact. Near-zero values in most settings confirm that edits successfully override model outputs.}
\label{tab:edit_success}
\end{table}

\begin{table*}
\centering
\setlength{\tabcolsep}{2pt}
\begin{tabular}{llccccccccc}
\toprule
\multirow{2}{*}{\textbf{Method}} & \multirow{2}{*}{\textbf{Model}} & \multirow{2}{*}{\textbf{Pruned}} & \multicolumn{2}{c}{\textbf{Train}} & \multicolumn{2}{c}{\textbf{Test}} & \multicolumn{3}{c}{\textbf{PPL (Wikitext-2) $\downarrow$}} \\
\cmidrule(lr){4-5} \cmidrule(lr){6-7} \cmidrule(lr){8-10}
 & & & \textbf{RSR $\uparrow$} & \textbf{Top-1 $\uparrow$} & \textbf{RSR $\uparrow$} & \textbf{Top-1 $\uparrow$} & \textbf{$M$} & \textbf{$M_e$} & \textbf{$M_p$} \\
\midrule
\multirow{3}{*}{ROME} & GPT-2 XL & 10.0\% & 83\% & 78\% & 82\% & 77\% & 17.80 & $44.51_{\pm 6.8}$ & 25.68$_{\pm 0.4}$ \\
 & LLaMA-3.2 & 10.0\% & 90\% & 75\% & 79\% & 72\% & 9.46 & 9.59$_{\pm 0.0}$ & 10.14$_{\pm 0.0}$ \\
& Qwen-2.5 & 15.3\% & 87\% & 77\% & 68\% & 66\% & 7.43 & 8.07$_{\pm 0.37}$ & 7.76$_{\pm 0.01}$ \\
\midrule
\multirow{3}{*}{MEMIT} & GPT-2 XL & 4.5\% & 82\% & 81\% & 74\% & 78\% & 17.80 & 17.90 & 19.00 \\
 & LLaMA-3.2 & 8.8\% & 87\% & 67\% & 78\% & 65\% & 9.46 & 10.78 & 12.53 \\
 & Qwen-2.5 & 7.9\% & 88\% & 67\% & 70\% & 60\% & 7.43 & 7.66 & 7.74 \\
\bottomrule
\end{tabular}
\caption{\textbf{Evaluation of edit reversal via the shared mask.} We report reversal performance (RSR and Top-1 Overlap) alongside model perplexity (PPL) for the original ($M$), edited ($M_e$), and pruned ($M_p$) models.}
\label{tab:reversal_detailed}
\end{table*}

\subsection{Results}
\paragraph{Editing Performance.}
Before analyzing reversal performance, we verify that editing succeeds: \autoref{tab:edit_success} 
confirms that the edited models strongly prefer the new object over the original one, with RSR and Top-1 Overlap near zero in most cases.
\paragraph{Reversal Performance.} \autoref{tab:reversal_detailed} shows that a single mask successfully reverses the majority of edits across both methods and architectures. For ROME on GPT-2 XL, pruning only 10\% of the edited layer's weights achieves a Reversal Success Rate (RSR) of 83\% on training set edits, with 78\% Top-1 Overlap indicating that the pruned model's top prediction matches the original model in most cases. MEMIT requires even less intervention: pruning just 4.5\% of the single edited layer's weights (0.9\% of total edited weights) yields 82\% RSR with 81\% Top-1 Overlap.

Crucially, these masks generalize to unseen edits. On the held-out test set, the ROME mask maintains 82\% RSR and 77\% Top-1 Overlap; the MEMIT mask achieves 74\% RSR with 78\% Top-1 Overlap. For Qwen-2.5, both ROME and MEMIT masks generalize to unseen edits, though ROME requires a slightly higher pruning rate (15.3\%) on this larger architecture. The consistency across all three models suggests that the shared mechanism is a general property of ROME and MEMIT rather than an artifact of a specific architecture.
\paragraph{Perplexity After Pruning.} 
Beyond reversal performance, we verify that our masks do not degrade general language modeling. \autoref{tab:reversal_detailed} shows perplexity on WikiText-2 for the original ($M$), edited ($M_e$), and pruned ($M_p$) models.

ROME edits can substantially harm model performance. On GPT-2 XL, perplexity increases from 17.8 to 44.5 after editing yielding a 2.5$\times$ degradation. Applying the mask reduces perplexity to 25.7, recovering much of the lost performance without modifying the edited weights. In extreme cases, ROME edits cause model collapse \cite{yang-etal-2024-butterfly} with a perplexity score exceeding one million. Our mask reduces these closer to the baseline levels (cf. \autoref{sec:appendix_a}). Qwen-2.5 shows a similar pattern: perplexity rises slightly from 7.43 to 8.07 after editing, and applying the mask brings it back down to 7.76, recovering most of the degradation.

LLaMA-3.2 proves more robust to ROME edits, with perplexity rising only from 9.46 to 9.59. The mask introduces minimal additional degradation (10.14), confirming that our intervention is targeted rather than destructive.

MEMIT tells a different story. Its edits barely affect perplexity on all three models (17.8 → 17.9 for GPT-2 XL; 9.46 → 10.78 for LLaMA-3.2), yet our mask still reverses over 80\% of edits by pruning under 9\% of weights without substantial damage to the performance. This indicates that MEMIT's edits, while less disruptive to general capabilities, still rely on a small subset of functionally critical weights to maintain the edit. Once these are suppressed, original knowledge resurfaces.

\section{Analysis}
\begin{table*}[h]
\centering
\resizebox{1.75\columnwidth}{!}{
\begin{tabular}{lrcccc}
\toprule
\textbf{Model} & \textbf{Method} & $M$ & $M_e$ & \textbf{$p$-value} & \textbf{Cohen's $d$}\\
\midrule
\multirow{2}{*}{GPT-2 XL} & ROME & $0.045_{\pm 0.11}$ & $0.866_{\pm 0.19}$ & $2.9\text{e-}165$ & $3.78$ \\
& MEMIT & $0.046_{\pm 0.11}$ & $0.614_{\pm 0.36}$ & $3.5\text{e-}156$ & $1.54$\\
\midrule
\multirow{2}{*}{LLaMA-3.2} & ROME & $2.46\text{e-}5_{\pm 6\text{e-}5}$ & $1.96\text{e-}4_{\pm 5\text{e-}4}$ & $1.6\text{e-}68$ & $0.34$ \\
& MEMIT & $2.48\text{e-}5_{\pm 6\text{e-}5}$ & $1.17\text{e-}4_{\pm 3\text{e-}4}$ & $1.3\text{e-}73$ & $0.35$ \\
\bottomrule
\end{tabular}
}
\caption{\textbf{Statistical analysis of probabilities strength on 1,000 samples across models}. We compare the mean output probability of the original model ($M$) on original facts versus the edited model ($M_e$) on edited facts.}
\label{tab:signal_stats_combined}
\end{table*}

\begin{figure*}[h]
    \centering
    \begin{subfigure}{0.48\linewidth}
        \centering
        LLaMA-3.2 \\ \vspace{0.2em}
        \includegraphics[width=\linewidth]{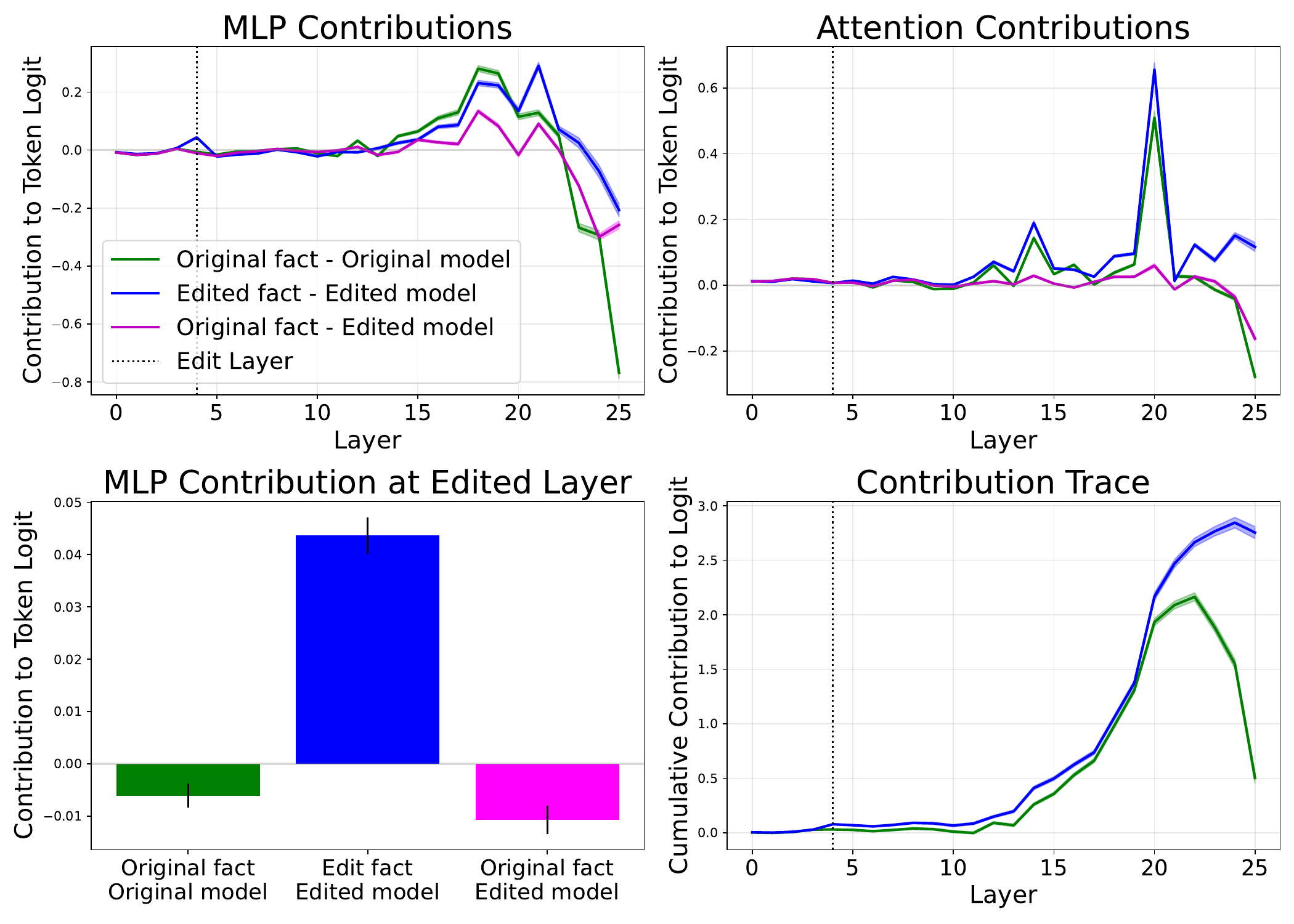}
    \end{subfigure}
    \hfill
    \begin{subfigure}{0.48\linewidth}
        \centering
        GPT-2 XL \\ \vspace{0.2em}
        \includegraphics[width=\linewidth]{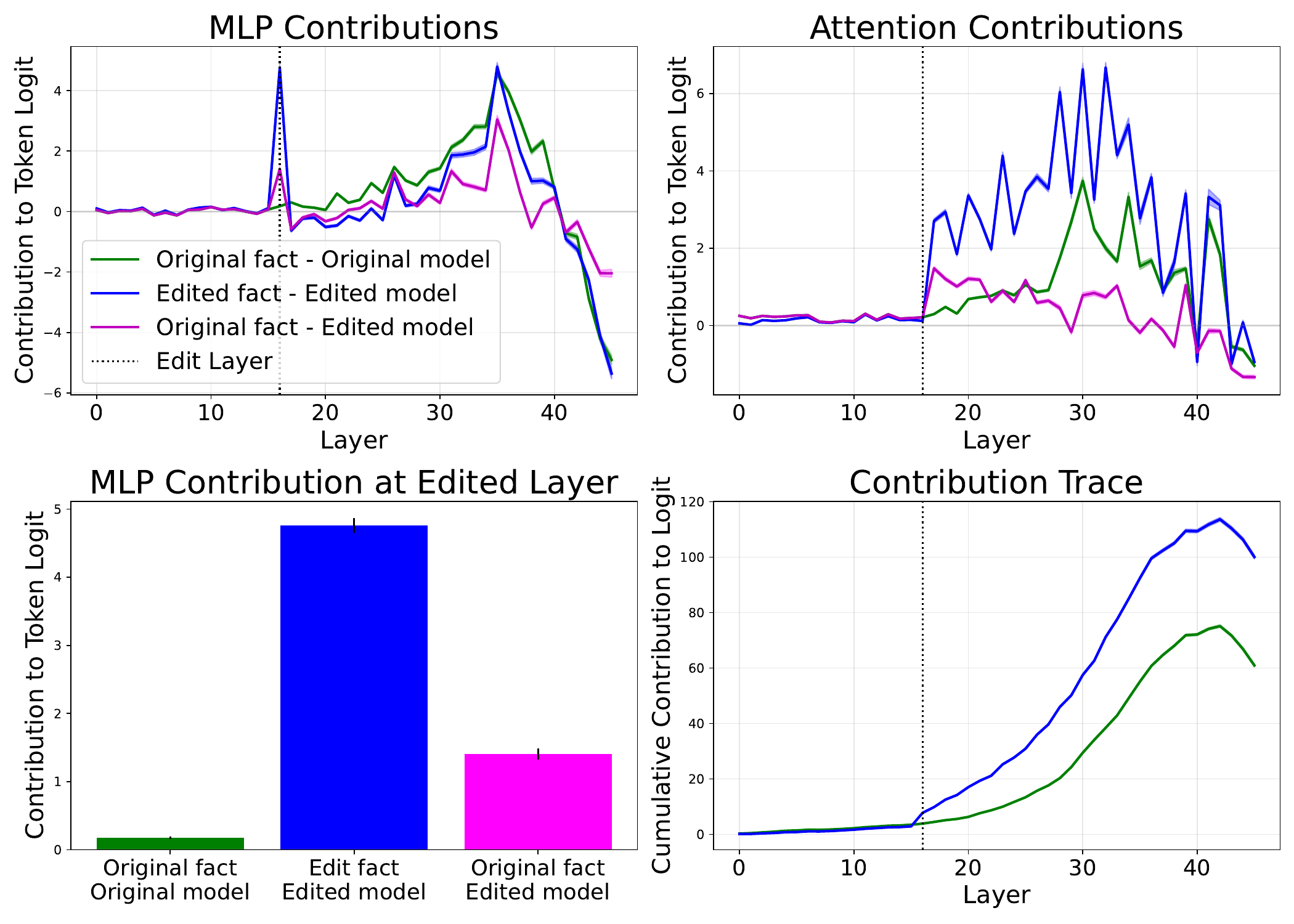}
    \end{subfigure}

    \caption{\textbf{Decomposition of residual stream} (mean and standard error over 1,000 samples) across LLaMA-3.2 and GPT-2 XL. 
    For each model: 
    Top Left: MLP contributions per layer. 
    Top Right: Attention contributions per layer. 
    Bottom Left: Comparison of MLP contributions at the edited layer. 
    Bottom Right: Overall logit trace across layers. Edits amplify signals at the edited layer, causing downstream attention spikes while MLP pathways continue to follow broadly similar trajectories for original and edited facts.}
    \label{fig:decomposition_comparison}
\end{figure*}

The reversal results suggest that a single mask can reverse diverse edits across methods and architectures by targeting a shared mechanism anchored in the edited weights. In this section, we analyze \emph{what constitutes this mechanism} by answering two questions: 1) How do ROME and MEMIT alter information flow within the model?; 2) What exactly does the trained mask target to reverse edits? We limit our analysis to GPT-2 XL and LLaMA-3.2.

\paragraph{Edits induce overattention.}
Both ROME and MEMIT force edited models to assign dramatically higher probabilities to edited facts. As shown in \autoref{tab:signal_stats_combined}, ROME on GPT-2 XL increases mean probability from 0.045 to 0.87 (Cohen's $d = 3.78$); LLaMA-3.2 shows the same pattern at lower absolute values ($d = 0.34$). These artificially elevated probabilities suggest that editing fundamentally alters information flow through the model. If knowledge can be retrieved through multiple pathways \cite{mcgrath2023hydraeffectemergentselfrepair, hase2023does}, \emph{how does editing succeed in producing such dominant output probabilities?} 

Recent work has documented attention-related phenomena in edited models: excessive attention to edited entities~\citep{wang2025revealing} and later-layer attention modules that cause the residual stream to revert toward original knowledge~\citep{xie-etal-2025-revealing-deceptiveness}. While previous work treats overattention and retention of original knowledge as side effects , we hypothesize that this \textbf{overattention is not merely a side effect but the mechanism by which edits succeed}: amplified signals hijack downstream attention, suppressing original facts without erasing them. To test this, we decompose the residual stream into MLP and attention contributions using the Logit Lens method~\citep{nostalgebraist2020interpreting}, measuring each component's contribution to the target token's logit (cf. \autoref{fig:decomposition_comparison}).

The two architectures differ in magnitude but share the same functional pattern. In GPT-2 XL, ROME injects a signal $35\times$ larger than baseline at the edited layer (0.13 vs 4.55), producing immediate attention spikes downstream. LLaMA-3.2 shows modest amplification at the edited layer, but attention contributions increase sharply in later layers (19-21) causing a delayed overattention effect. The critical observation is that \emph{MLP contributions beyond the edited layer follow broadly similar trajectories for edited and original facts} in both architectures. Downstream MLPs continue processing original knowledge; the main divergence occurs in the attention layers. The cumulative logit traces confirm this shared mechanism. In both models, edited facts accumulate substantially higher contributions than original facts, with the gap widening through downstream layers. MEMIT exhibits the same pattern despite distributing edits across several layers (cf. \autoref{sec:appendix_b}).

These observations indicate that overattention is central to how edits produce dominant outputs. However, a key question remains: do different edits induce overattention through independent pathways, or do they share a common mechanism anchored in the edited weights? Is amplified attention a side-effect of editing or its core mechanism? 

\begin{figure*}
    \centering
    \begin{subfigure}{0.48\linewidth}
        \centering
        LLaMA-3.2 \\ \vspace{0.2em}
        \includegraphics[width=\linewidth]{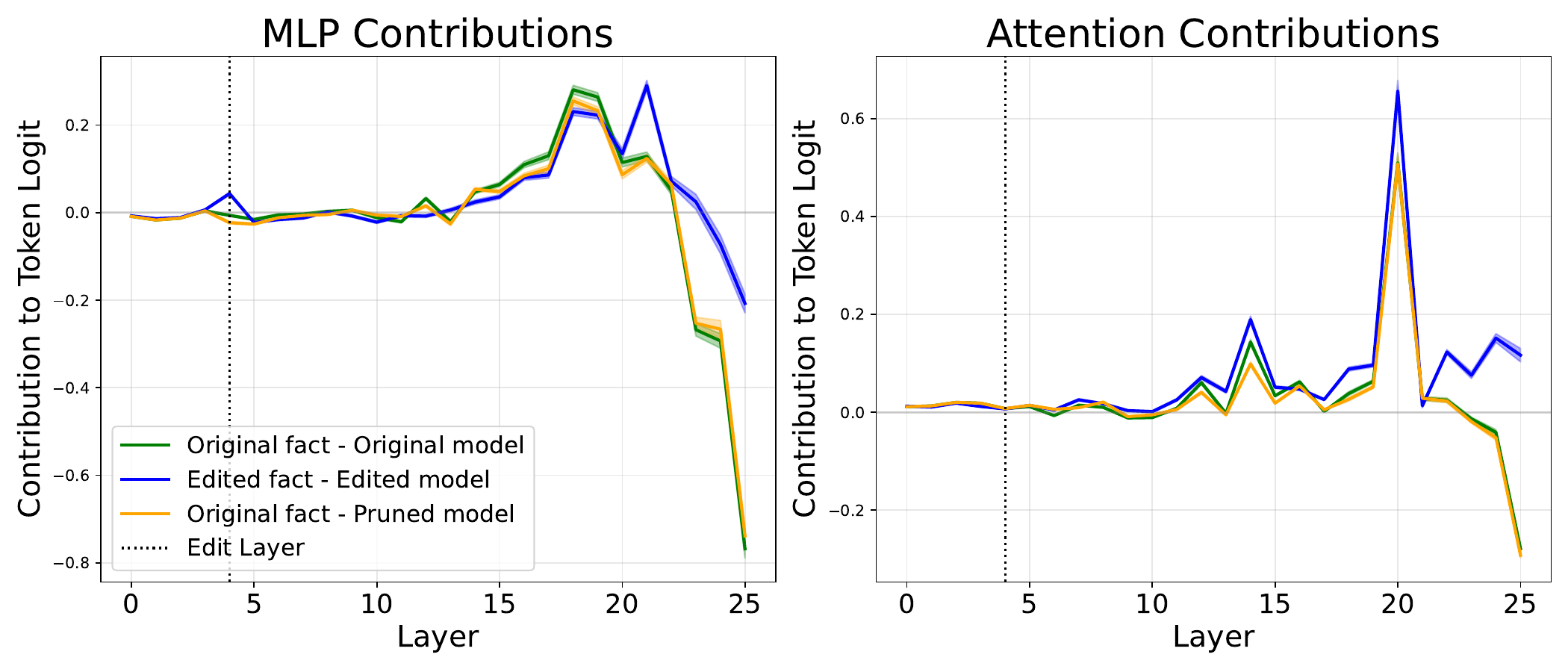}
    \end{subfigure}
    \hfill
    \begin{subfigure}{0.48\linewidth}
        \centering
        GPT-2 XL \\ \vspace{0.2em}
        \includegraphics[width=\linewidth]{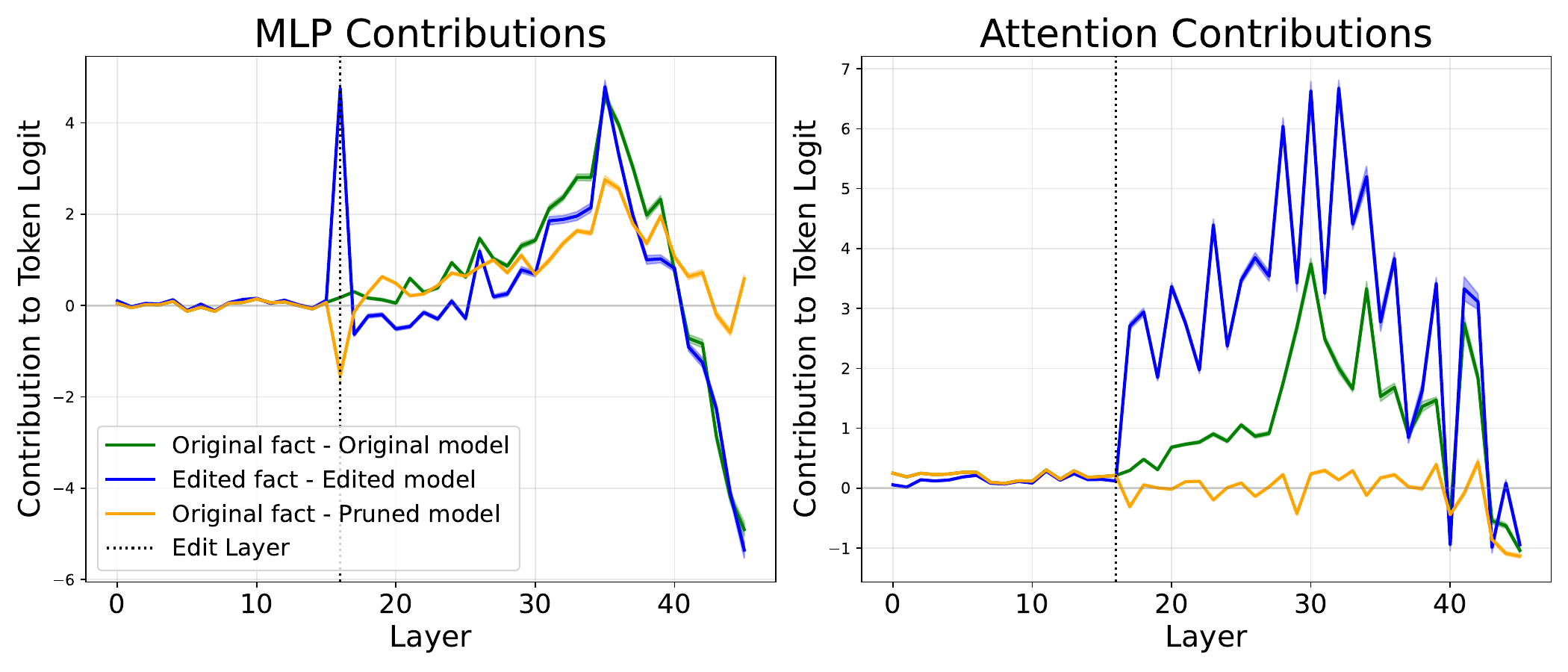}
    \end{subfigure}
    \caption{\textbf{Decomposition of residual stream for original, edited and pruned models} (mean and standard error over 1,000 samples) across both models.}
    \label{fig:decomposition_pruned_comparison}
\end{figure*}
\paragraph{Mask eliminates overattention.} 
The mask's high reversal performance on unseen edits (cf. \autoref{tab:reversal_detailed}) suggests that a shared mechanism critical to maintaining edits exists. To identify what the mask targets, we decompose the residual stream of the pruned model (cf. \autoref{fig:decomposition_pruned_comparison}) and compare it with the edited and original models from \autoref{fig:decomposition_comparison}.

In GPT-2 XL, the MLP spike at the edited layer is eliminated. Critically, attention contributions in downstream layers are substantially reduced, while MLP contributions remain largely unaffected, following trajectories similar to the original model.

LLaMA-3.2 shows the same pattern. The late-layer attention spikes (layers 19-21) that dominate in the edited model are fully eliminated in the pruned model, and the subsequent MLP spike (layers 20-22) disappears along with it. This confirms that the late MLP activity was a consequence of overattention.

\begin{figure}[t]
    \centering
    Original \hspace{3.4em} Edited \hspace{3em} Pruned \\
    \includegraphics[width=\linewidth]{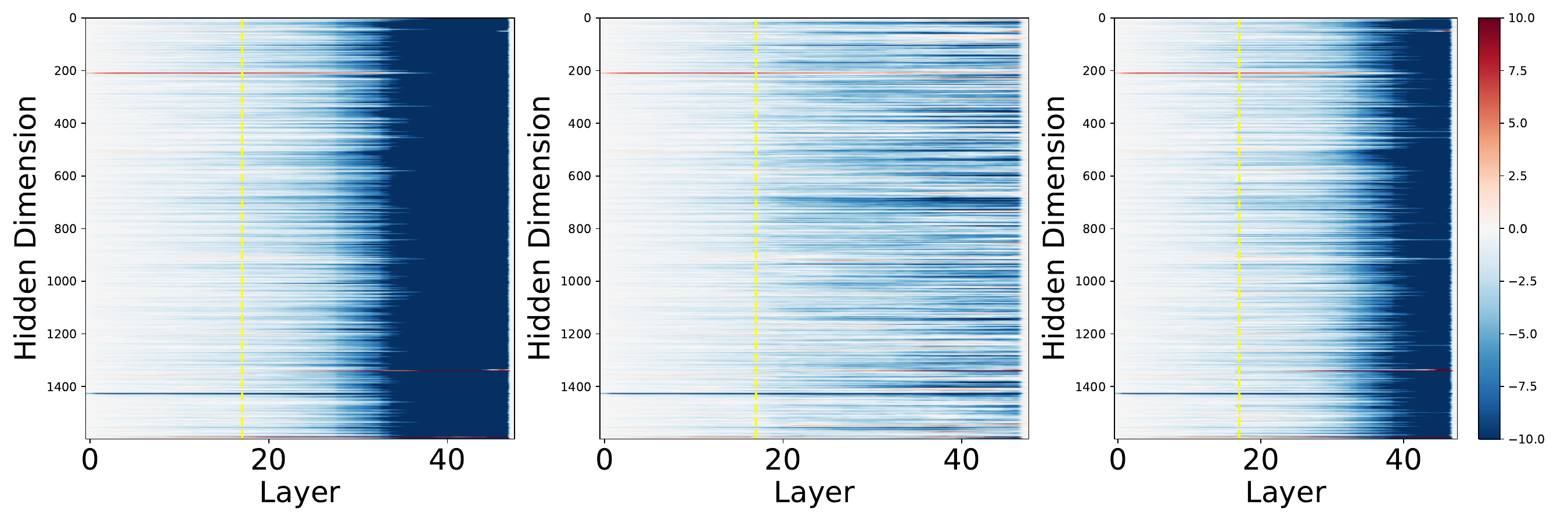}
    \vspace{1mm}
    \includegraphics[width=\linewidth]{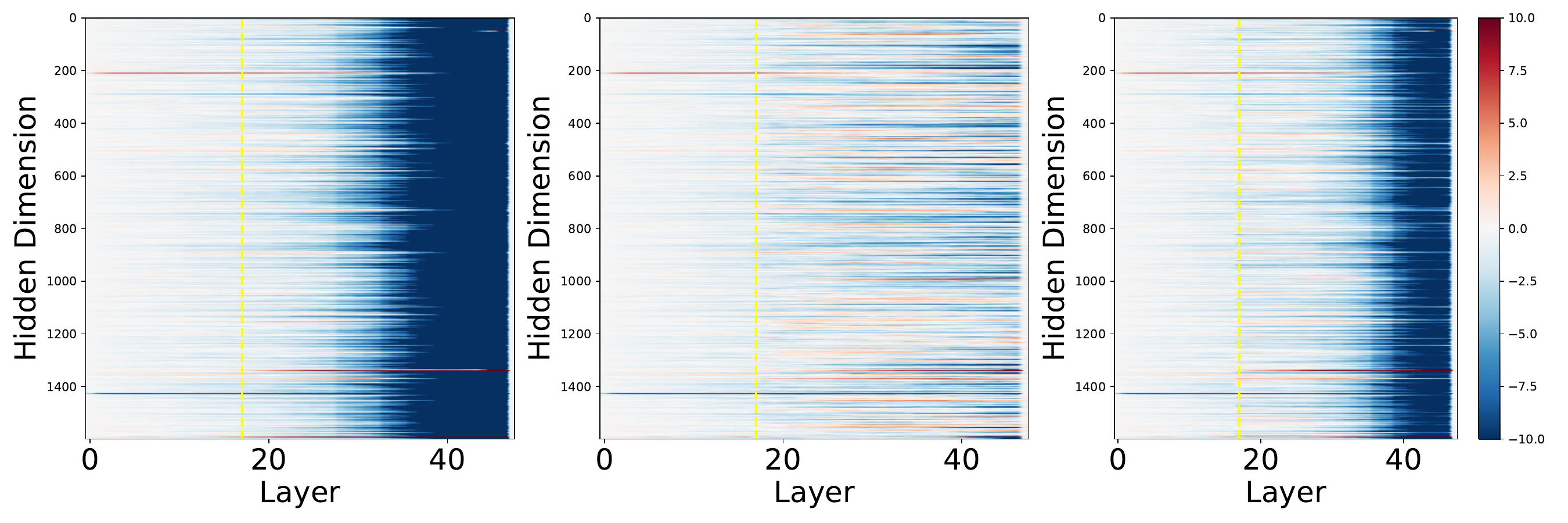}

    \caption{\textbf{Activations across hidden dimensions and layers.} Residual stream activations for original (left), edited (middle), and pruned (right) GPT-2 XL models. Yellow dashed line marks the edited layer.}
    \label{fig:activ_recovery}
\end{figure}
\textbf{In both architectures, the masks trained on semantically diverse edits converge to a common solution: eliminating overattention while preserving MLP pathways.} The mask trained to restore original knowledge independently discovers that targeting overattention is necessary and sufficient for reversal. The fact that the same weight positions are masked across different edits suggests a shared update pattern in how ROME and MEMIT induce overattention. This convergence is direct evidence that overattention is not a side effect but the shared structural mechanism that ROME and MEMIT exploit. Additionally, \autoref{fig:activ_recovery} illustrates this restoration: the pruned model recovers activation patterns closely resembling the original, confirming that the mask reverses the edit's effect on the residual stream. Further analysis of mask structure and other pruning experiments can be found in \autoref{sec:appendix_c}.

\paragraph{Mask blocks new edits.}
The mask's ability to reverse edits demonstrates that targeting the shared subspace is sufficient for reversal. But is this subspace also \textit{necessary} for edits to succeed or can it be avoided by taking other computational pathways inside the model? To test this, we inject the learned mask into the forward pass \textit{during} the editing process itself.

\begin{figure}
    \centering
    \includegraphics[width=0.95\linewidth]{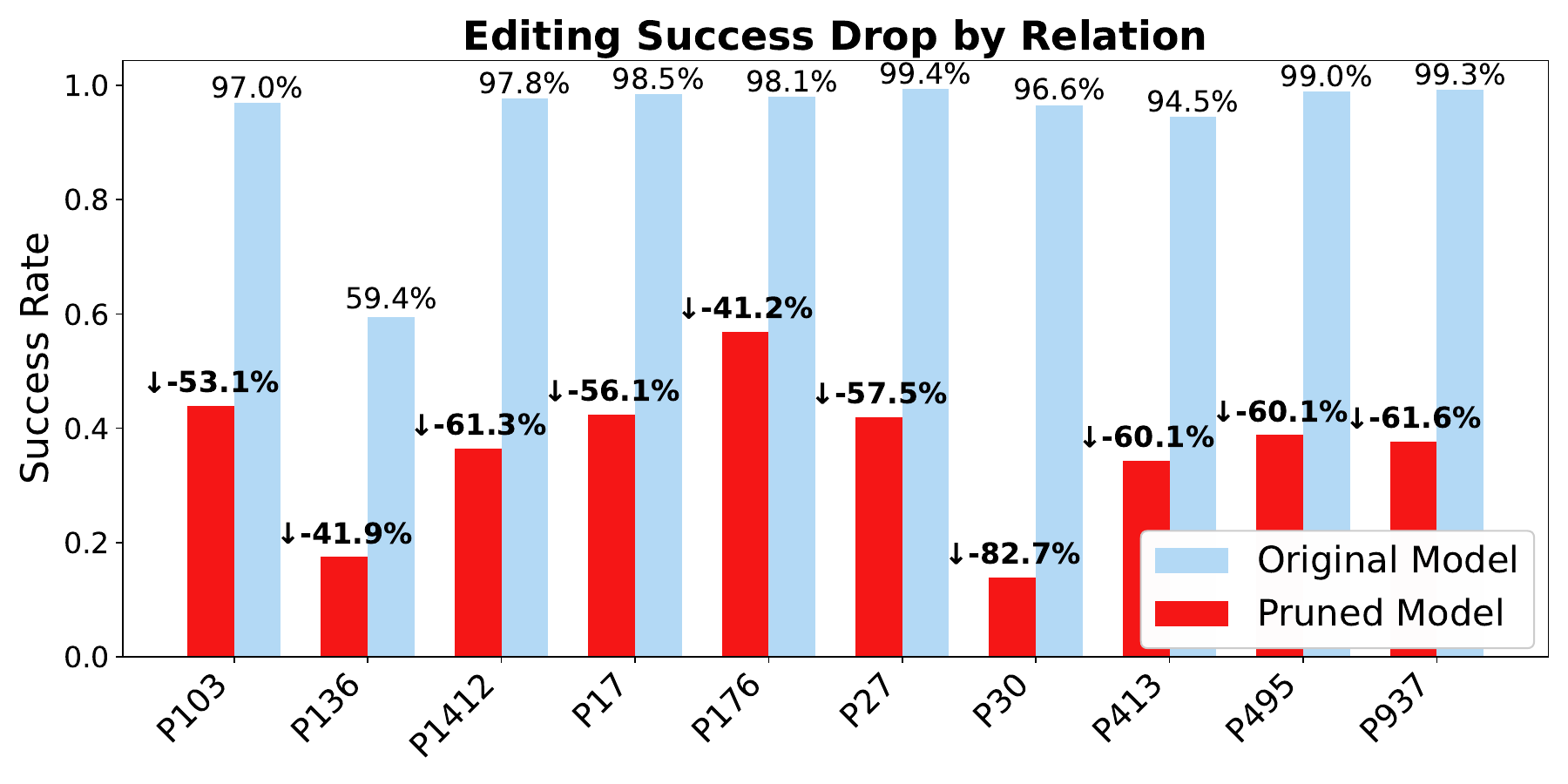}
    \caption{
        \textbf{Editing Success Drop.}
        Comparison of standard ROME editing efficacy (blue bars) versus editing the model while the shared mask is active (red bars).
    }
    \label{fig:es_drop}
\end{figure}

If the subspace targeted by the mask is merely associated with successful edits but not required for them, editing should succeed through alternative pathways. Instead,  as shown in~\autoref{fig:es_drop}, editing success rate drops from 98\% to 38\%. This drop in editing success is consistent across relation types, indicating that the mask does not target relation-specific structure but a general structure. This result confirms that the weight subspace identified by our mask is not incidental to editing --- it is structurally necessary. ROME struggles to bypass this shared mechanism to inject new facts.

\section{Discussion}
Our findings have several implications for understanding knowledge editing and for AI safety. 
\paragraph{Overattention as mechanism.}Prior work documents overattention in edited models as a problem: a source of specificity failure where edits affect unrelated knowledge~\citep{wang2025revealing}, or a reason an edited model reverts to original facts under adversarial prompts~\citep{xie-etal-2025-revealing-deceptiveness}. Our findings reframe overattention as the core mechanism that makes editing work. A single mask recovers over 80\% of training edits and 70\% of unseen edits (cf.~\autoref{tab:reversal_detailed}) by targeting the same weight positions across semantically diverse facts. This would be impossible if edits introduced only fact-specific updates without a shared mechanism critical for maintaining them. The mask converges on eliminating overattention --- the shared target that sustains edits across diverse facts. Moreover, injecting the mask during the editing process confirms that ROME cannot route around the blocked weights. This demonstrates that overattention is not a side effect but the structural basis that ROME and MEMIT make use of to edit facts.

\paragraph{Knowledge is not erased.}On a conceptual level, these findings challenge the current knowledge editing paradigm. Knowledge is densely interconnected: a single fact relates to thousands of others. What does modifying a factual association actually mean? In a rigorous sense, it should keep the model's knowledge base consistent: modifying one fact should propagate to related associations, spanning a coherent counterfactual world. If we change the capital of France from Paris to Lyon, the model should reflect an alternative reality where this is true across all related queries. Current editing methods fail to achieve this. They do not propagate edits to related knowledge, creating ripple effects~\citep{li-etal-2023-evaluating-dependencies, cohen-etal-2024-evaluating, qin-etal-2024-new, hsueh-etal-2024-editing} that distort both related and unrelated facts. Our findings suggest why: ROME and MEMIT do not modify the knowledge graph - they hijack attention to suppress its retrieval. The original facts remain encoded; overattention simply prevents them from surfacing. This explains both why edits fail to propagate and why they can be reversed by targeting a small shared subspace.

\paragraph{Fundamental limitations of locate-and-edit methods.}If ROME and MEMIT succeed by hijacking attention rather than modifying stored knowledge, this points to a fundamental limitation of the ROME/MEMIT paradigm that may extend to other locate-and-edit methods. The associative memory hypothesis~\citep{geva-etal-2021-transformer} motivating ROME and MEMIT may be incomplete: even if facts are stored in MLP layers, retrieval involves attention, and current methods exploit this dependency rather than updating the stored associations. A shared functional subspace --- one that generalizes across semantically diverse edits --- may suggest an upper bound on what locate-and-edit methods can achieve. They may be inherently limited to suppressing knowledge rather than modifying it.

\paragraph{Implications for AI safety.}On the defensive side, edits are more reversible than previously assumed: a sparse mask trained on a small set of edits can recover original knowledge across unseen facts. The same mask can also lock the model against future edits, as our blocking experiment demonstrates. Prior work has shown that edited facts are detectable from internal representations~\citep{youssef-etal-2025-fact, youssef2025tracingreversingrankonemodel}. Our findings may explain why: if all edits exploit a shared mechanism that produces distinctive attention patterns, they leave a common signature that classifiers can learn to recognize. Overall, understanding this shared mechanism brings us closer to defending against unwanted edits of open-weight models, by both locking models against future interventions or reversing the existing edits.
\section{Conclusion}
We demonstrated that locate-and-edit methods exploit a shared mechanism to maintain edits across semantically diverse facts. A single sparse mask generalizes to unseen edits with over 70\% success rate, revealing that the majority of edits rely on the same functional structure. Residual stream decomposition shows that the mask converges on eliminating overattention, while preserving MLP pathways that continue to encode the original knowledge. Injecting the mask during editing drops success from 98\% to 38\%. This confirms that this mechanism is not merely sufficient for reversal, but also necessary for edits to succeed.

These findings highlight a fundamental limitation of locate-and-edit methods: rather than modifying stored knowledge, these methods suppress its retrieval. 
Our analysis enables both detection and defense against unwanted edits --- the same mask that reverses edits can also block edits, offering a practical way to lock models against malicious edits. 
Future work should extend this analysis to other locate-and-edit methods such as AlphaEdit~\cite{fang2025alphaedit} and meta-learning approaches ~\cite{mitchell2022fast, tan2023massive}.

\section*{Limitations}
In this work, we focused on locate-and-edit KEs like ROME and MEMIT because of their widespread use and computational efficiency. Meta-learning KEs also adapt the model's parameters, and might be changing facts retrieval in LLMs in a different way. We did not consider these methods in our work because of the high computational costs associated with training editing hypernetworks. 

\bibliography{custom}

@inproceedings{youssef-etal-2024-queen,
    title = "{The Queen of England is not England's Queen: On the Lack of Factual Coherency in PLMs}",
    author = {Youssef, Paul  and
      Schl{\"o}tterer, J{\"o}rg  and
      Seifert, Christin},
    editor = "Graham, Yvette  and
      Purver, Matthew",
    booktitle = "Findings of the Association for Computational Linguistics: EACL 2024",
    month = mar,
    year = "2024",
    address = "St. Julian{'}s, Malta",
    publisher = "Association for Computational Linguistics",
    url = "https://aclanthology.org/2024.findings-eacl.155/",
    doi = "10.18653/v1/2024.findings-eacl.155",
    pages = "2342--2354"
}

@misc{qwen2025qwen25technicalreport,
      title={{Qwen2.5 Technical Report}}, 
      author={Qwen and : and An Yang and Baosong Yang and Beichen Zhang and Binyuan Hui and Bo Zheng and Bowen Yu and Chengyuan Li and Dayiheng Liu and Fei Huang and Haoran Wei and Huan Lin and Jian Yang and Jianhong Tu and Jianwei Zhang and Jianxin Yang and Jiaxi Yang and Jingren Zhou and Junyang Lin and Kai Dang and Keming Lu and Keqin Bao and Kexin Yang and Le Yu and Mei Li and Mingfeng Xue and Pei Zhang and Qin Zhu and Rui Men and Runji Lin and Tianhao Li and Tianyi Tang and Tingyu Xia and Xingzhang Ren and Xuancheng Ren and Yang Fan and Yang Su and Yichang Zhang and Yu Wan and Yuqiong Liu and Zeyu Cui and Zhenru Zhang and Zihan Qiu},
      year={2025},
      eprint={2412.15115},
      archivePrefix={arXiv},
      primaryClass={cs.CL},
      url={https://arxiv.org/abs/2412.15115}, 
}

@inproceedings{
wang2025revealing,
title={Revealing and Mitigating Over-Attention in Knowledge Editing},
author={Pinzheng Wang and Zecheng Tang and Keyan Zhou and Juntao Li and Qiaoming Zhu and Min Zhang},
booktitle={The Thirteenth International Conference on Learning Representations},
year={2025},
url={https://openreview.net/forum?id=4l3AH8Bhmt}
}

@inproceedings{xie-etal-2025-revealing-deceptiveness,
    title = "Revealing the Deceptiveness of Knowledge Editing: A Mechanistic Analysis of Superficial Editing",
    author = "Xie, Jiakuan  and
      Cao, Pengfei  and
      Chen, Yubo  and
      Liu, Kang  and
      Zhao, Jun",
    editor = "Che, Wanxiang  and
      Nabende, Joyce  and
      Shutova, Ekaterina  and
      Pilehvar, Mohammad Taher",
    booktitle = "Proceedings of the 63rd Annual Meeting of the Association for Computational Linguistics (Volume 1: Long Papers)",
    month = jul,
    year = "2025",
    address = "Vienna, Austria",
    publisher = "Association for Computational Linguistics",
    url = "https://aclanthology.org/2025.acl-long.868/",
    doi = "10.18653/v1/2025.acl-long.868",
    pages = "17756--17780",
    ISBN = "979-8-89176-251-0"
}

@inproceedings{guo_calibration,
author = {Guo, Chuan and Pleiss, Geoff and Sun, Yu and Weinberger, Kilian Q.},
title = {On calibration of modern neural networks},
year = {2017},
publisher = {JMLR.org},
booktitle = {Proceedings of the 34th International Conference on Machine Learning - Volume 70},
pages = {1321–1330},
numpages = {10},
location = {Sydney, NSW, Australia},
series = {ICML'17}
}

@article{cohen-etal-2024-evaluating,
    title = "Evaluating the Ripple Effects of Knowledge Editing in Language Models",
    author = "Cohen, Roi  and
      Biran, Eden  and
      Yoran, Ori  and
      Globerson, Amir  and
      Geva, Mor",
    journal = "Transactions of the Association for Computational Linguistics",
    volume = "12",
    year = "2024",
    address = "Cambridge, MA",
    publisher = "MIT Press",
    url = "https://aclanthology.org/2024.tacl-1.16/",
    doi = "10.1162/tacl_a_00644",
    pages = "283--298",
}

@inproceedings{li-etal-2023-evaluating-dependencies,
    title = "Evaluating Dependencies in Fact Editing for Language Models: Specificity and Implication Awareness",
    author = "Li, Zichao  and
      Arous, Ines  and
      Reddy, Siva  and
      Cheung, Jackie",
    editor = "Bouamor, Houda  and
      Pino, Juan  and
      Bali, Kalika",
    booktitle = "Findings of the Association for Computational Linguistics: EMNLP 2023",
    month = dec,
    year = "2023",
    address = "Singapore",
    publisher = "Association for Computational Linguistics",
    url = "https://aclanthology.org/2023.findings-emnlp.511/",
    doi = "10.18653/v1/2023.findings-emnlp.511",
    pages = "7623--7636",
}

@inproceedings{qin-etal-2024-new,
    title = "Why Does New Knowledge Create Messy Ripple Effects in {LLM}s?",
    author = "Qin, Jiaxin  and
      Zhang, Zixuan  and
      Han, Chi  and
      Yu, Pengfei  and
      Li, Manling  and
      Ji, Heng",
    editor = "Al-Onaizan, Yaser  and
      Bansal, Mohit  and
      Chen, Yun-Nung",
    booktitle = "Proceedings of the 2024 Conference on Empirical Methods in Natural Language Processing",
    month = nov,
    year = "2024",
    address = "Miami, Florida, USA",
    publisher = "Association for Computational Linguistics",
    url = "https://aclanthology.org/2024.emnlp-main.700/",
    doi = "10.18653/v1/2024.emnlp-main.700",
    pages = "12602--12609",
}

@inproceedings{
fang2025alphaedit,
title={AlphaEdit: Null-Space Constrained Model Editing for Language Models},
author={Junfeng Fang and Houcheng Jiang and Kun Wang and Yunshan Ma and Jie Shi and Xiang Wang and Xiangnan He and Tat-Seng Chua},
booktitle={The Thirteenth International Conference on Learning Representations},
year={2025},
url={https://openreview.net/forum?id=HvSytvg3Jh}
}

@inproceedings{bayazit-etal-2024-discovering,
    title = "Discovering Knowledge-Critical Subnetworks in Pretrained Language Models",
    author = "Bayazit, Deniz  and
      Foroutan, Negar  and
      Chen, Zeming  and
      Weiss, Gail  and
      Bosselut, Antoine",
    editor = "Al-Onaizan, Yaser  and
      Bansal, Mohit  and
      Chen, Yun-Nung",
    booktitle = "Proceedings of the 2024 Conference on Empirical Methods in Natural Language Processing",
    month = nov,
    year = "2024",
    address = "Miami, Florida, USA",
    publisher = "Association for Computational Linguistics",
    url = "https://aclanthology.org/2024.emnlp-main.376/",
    doi = "10.18653/v1/2024.emnlp-main.376",
    pages = "6549--6583"
}

@inproceedings{
  louizos2018learning,
  title={Learning Sparse Neural Networks through L0 Regularization},
  author={Christos Louizos and Max Welling and Diederik P. Kingma},
  booktitle={International Conference on Learning Representations},
  year={2018},
  url={https://openreview.net/forum?id=H1Y8hhg0b},
}

@inproceedings{
maddison2017the,
title={The Concrete Distribution: A Continuous Relaxation of Discrete Random Variables},
author={Chris J. Maddison and Andriy Mnih and Yee Whye Teh},
booktitle={International Conference on Learning Representations},
year={2017},
url={https://openreview.net/forum?id=S1jE5L5gl}
}

@inproceedings{hsueh-etal-2024-editing,
    title = "Editing the Mind of Giants: An In-Depth Exploration of Pitfalls of Knowledge Editing in Large Language Models",
    author = "Hsueh, Cheng-Hsun  and
      Huang, Paul Kuo-Ming  and
      Lin, Tzu-Han  and
      Liao, Che Wei  and
      Fang, Hung-Chieh  and
      Huang, Chao-Wei  and
      Chen, Yun-Nung",
    editor = "Al-Onaizan, Yaser  and
      Bansal, Mohit  and
      Chen, Yun-Nung",
    booktitle = "Findings of the Association for Computational Linguistics: EMNLP 2024",
    month = nov,
    year = "2024",
    address = "Miami, Florida, USA",
    publisher = "Association for Computational Linguistics",
    url = "https://aclanthology.org/2024.findings-emnlp.550/",
    doi = "10.18653/v1/2024.findings-emnlp.550",
    pages = "9417--9429"
}

@inproceedings{
meng2022locating,
title={Locating and Editing Factual Associations in {GPT}},
author={Kevin Meng and David Bau and Alex J Andonian and Yonatan Belinkov},
booktitle={Advances in Neural Information Processing Systems},
editor={Alice H. Oh and Alekh Agarwal and Danielle Belgrave and Kyunghyun Cho},
year={2022},
url={https://openreview.net/forum?id=-h6WAS6eE4}
}

@inproceedings{
meng2023massediting,
title={Mass-Editing Memory in a Transformer},
author={Kevin Meng and Arnab Sen Sharma and Alex J Andonian and Yonatan Belinkov and David Bau},
booktitle={The Eleventh International Conference on Learning Representations },
year={2023},
url={https://openreview.net/forum?id=MkbcAHIYgyS}
}

@inproceedings{youssef-etal-2026-persuasion,
    title = "{Persuasion Tokens for Editing Factual Knowledge in LLMs}",
    author = {Youssef, Paul  and
      Seifert, Christin  and
      Schl{\"o}tterer, J{\"o}rg},
    editor = "Demberg, Vera  and
      Inui, Kentaro  and
      Marquez, Llu{\'i}s",
    booktitle = "Proceedings of the 19th Conference of the {E}uropean Chapter of the {A}ssociation for {C}omputational {L}inguistics (Volume 2: Short Papers)",
    month = mar,
    year = "2026",
    address = "Rabat, Morocco",
    publisher = "Association for Computational Linguistics",
    url = "https://aclanthology.org/2026.eacl-short.35/",
    doi = "10.18653/v1/2026.eacl-short.35",
    pages = "475--486",
    ISBN = "979-8-89176-381-4",
}

@inproceedings{
youssef2025tracingreversingrankonemodel,
title={{Tracing and Reversing Edits in LLMs}},
author={Paul Youssef and Zhixue Zhao and Christin Seifert and J{\"o}rg Schl{\"o}tterer},
booktitle={The Fourteenth International Conference on Learning Representations},
year={2026},
url={https://openreview.net/forum?id=AiT8F6pbfi}
}

@inproceedings{petroni-etal-2019-language,
    title = "Language Models as Knowledge Bases?",
    author = {Petroni, Fabio  and
      Rockt{\"a}schel, Tim  and
      Riedel, Sebastian  and
      Lewis, Patrick  and
      Bakhtin, Anton  and
      Wu, Yuxiang  and
      Miller, Alexander},
    editor = "Inui, Kentaro  and
      Jiang, Jing  and
      Ng, Vincent  and
      Wan, Xiaojun",
    booktitle = "Proceedings of the 2019 Conference on Empirical Methods in Natural Language Processing and the 9th International Joint Conference on Natural Language Processing (EMNLP-IJCNLP)",
    month = nov,
    year = "2019",
    address = "Hong Kong, China",
    publisher = "Association for Computational Linguistics",
    url = "https://aclanthology.org/D19-1250/",
    doi = "10.18653/v1/D19-1250",
    pages = "2463--2473",
}

@inproceedings{youssef-etal-2023-give,
    title = "{Give Me the Facts! A Survey on Factual Knowledge Probing in Pre-trained Language Models}",
    author = {Youssef, Paul  and
      Kora{\c{s}}, Osman  and
      Li, Meijie  and
      Schl{\"o}tterer, J{\"o}rg  and
      Seifert, Christin},
    editor = "Bouamor, Houda  and
      Pino, Juan  and
      Bali, Kalika",
    booktitle = "Findings of the Association for Computational Linguistics: EMNLP 2023",
    month = dec,
    year = "2023",
    address = "Singapore",
    publisher = "Association for Computational Linguistics",
    url = "https://aclanthology.org/2023.findings-emnlp.1043/",
    doi = "10.18653/v1/2023.findings-emnlp.1043",
    pages = "15588--15605"
}

@inproceedings{geva-etal-2021-transformer,
    title = "Transformer Feed-Forward Layers Are Key-Value Memories",
    author = "Geva, Mor  and
      Schuster, Roei  and
      Berant, Jonathan  and
      Levy, Omer",
    editor = "Moens, Marie-Francine  and
      Huang, Xuanjing  and
      Specia, Lucia  and
      Yih, Scott Wen-tau",
    booktitle = "Proceedings of the 2021 Conference on Empirical Methods in Natural Language Processing",
    month = nov,
    year = "2021",
    address = "Online and Punta Cana, Dominican Republic",
    publisher = "Association for Computational Linguistics",
    url = "https://aclanthology.org/2021.emnlp-main.446/",
    doi = "10.18653/v1/2021.emnlp-main.446",
    pages = "5484--5495",
}

@misc{li2024knowledgeeditinglargelanguage,
      title={Knowledge Editing for Large Language Model with Knowledge Neuronal Ensemble}, 
      author={Yongchang Li and Yujin Zhu and Tao Yan and Shijian Fan and Gang Wu and Liang Xu},
      year={2024},
      eprint={2412.20637},
      archivePrefix={arXiv},
      primaryClass={cs.CL},
      url={https://arxiv.org/abs/2412.20637}, 
}

@misc{mcgrath2023hydraeffectemergentselfrepair,
      title={The Hydra Effect: Emergent Self-repair in Language Model Computations}, 
      author={Thomas McGrath and Matthew Rahtz and Janos Kramar and Vladimir Mikulik and Shane Legg},
      year={2023},
      eprint={2307.15771},
      archivePrefix={arXiv},
      primaryClass={cs.LG},
      url={https://arxiv.org/abs/2307.15771}, 
}

@inproceedings{
hase2023does,
title={Does Localization Inform Editing? Surprising Differences in Causality-Based Localization vs. Knowledge Editing in Language Models},
author={Peter Hase and Mohit Bansal and Been Kim and Asma Ghandeharioun},
booktitle={Thirty-seventh Conference on Neural Information Processing Systems},
year={2023},
url={https://openreview.net/forum?id=EldbUlZtbd}
}

@inproceedings{Loshchilov2017DecoupledWD,
  title={Decoupled Weight Decay Regularization},
  author={Ilya Loshchilov and Frank Hutter},
  booktitle={International Conference on Learning Representations},
  year={2017},
  url={https://api.semanticscholar.org/CorpusID:53592270}
}

@inproceedings{wang-etal-2024-easyedit,
    title = "{E}asy{E}dit: An Easy-to-use Knowledge Editing Framework for Large Language Models",
    author = "Wang, Peng  and
      Zhang, Ningyu  and
      Tian, Bozhong  and
      Xi, Zekun  and
      Yao, Yunzhi  and
      Xu, Ziwen  and
      Wang, Mengru  and
      Mao, Shengyu  and
      Wang, Xiaohan  and
      Cheng, Siyuan  and
      Liu, Kangwei  and
      Ni, Yuansheng  and
      Zheng, Guozhou  and
      Chen, Huajun",
    editor = "Cao, Yixin  and
      Feng, Yang  and
      Xiong, Deyi",
    booktitle = "Proceedings of the 62nd Annual Meeting of the Association for Computational Linguistics (Volume 3: System Demonstrations)",
    month = aug,
    year = "2024",
    address = "Bangkok, Thailand",
    publisher = "Association for Computational Linguistics",
    url = "https://aclanthology.org/2024.acl-demos.9/",
    doi = "10.18653/v1/2024.acl-demos.9",
    pages = "82--93"
}

@inproceedings{yang-etal-2024-butterfly,
    title = "The Butterfly Effect of Model Editing: Few Edits Can Trigger Large Language Models Collapse",
    author = "Yang, Wanli  and
      Sun, Fei  and
      Ma, Xinyu  and
      Liu, Xun  and
      Yin, Dawei  and
      Cheng, Xueqi",
    editor = "Ku, Lun-Wei  and
      Martins, Andre  and
      Srikumar, Vivek",
    booktitle = "Findings of the Association for Computational Linguistics ACL 2024",
    month = aug,
    year = "2024",
    address = "Bangkok, Thailand and virtual meeting",
    publisher = "Association for Computational Linguistics",
    url = "https://aclanthology.org/2024.findings-acl.322",
    pages = "5419--5437",
}

@inproceedings{mitchell2022fast,
    title={Fast Model Editing at Scale},
    author={Eric Mitchell and Charles Lin and Antoine Bosselut and Chelsea Finn and Christopher D Manning},
    booktitle={International Conference on Learning Representations},
    year={2022},
    url={https://openreview.net/pdf?id=0DcZxeWfOPt}
}

@inproceedings{hartvigsen2023aging,
  title={Aging with GRACE: Lifelong Model Editing with Discrete Key-Value Adaptors},
  author={Hartvigsen, Thomas and Sankaranarayanan, Swami and Palangi, Hamid and Kim, Yoon and Ghassemi, Marzyeh},
  booktitle={Advances in Neural Information Processing Systems},
  year={2023}
}

@inproceedings{mitchell2022memory,
    title={Memory-Based Model Editing at Scale},
    author={Mitchell, Eric and Lin, Charles and Bosselut, Antoine and Finn, Chelsea and Manning, Christopher D.},
    booktitle={International Conference on Machine Learning},
    url={https://arxiv.org/pdf/2206.06520.pdf},
    year={2022},
}

@inproceedings{zheng-etal-2023-edit,
    title = "{Can We Edit Factual Knowledge by In-Context Learning?}",
    author = "Zheng, Ce  and
      Li, Lei  and
      Dong, Qingxiu  and
      Fan, Yuxuan  and
      Wu, Zhiyong  and
      Xu, Jingjing  and
      Chang, Baobao",
    editor = "Bouamor, Houda  and
      Pino, Juan  and
      Bali, Kalika",
    booktitle = "Proceedings of the 2023 Conference on Empirical Methods in Natural Language Processing",
    month = dec,
    year = "2023",
    address = "Singapore",
    publisher = "Association for Computational Linguistics",
    url = "https://aclanthology.org/2023.emnlp-main.296/",
    doi = "10.18653/v1/2023.emnlp-main.296",
    pages = "4862--4876"
}

@inproceedings{
merity2017pointer,
title={Pointer Sentinel Mixture Models},
author={Stephen Merity and Caiming Xiong and James Bradbury and Richard Socher},
booktitle={International Conference on Learning Representations},
year={2017},
url={https://openreview.net/forum?id=Byj72udxe}
}

@inproceedings{Radford2019LanguageMA,
  title={Language Models are Unsupervised Multitask Learners},
  author={Alec Radford and Jeff Wu and Rewon Child and David Luan and Dario Amodei and Ilya Sutskever},
  year={2019},
  url={https://api.semanticscholar.org/CorpusID:160025533}
}

@article{Huang2024, 
author = {Wei Huang and Xingyu Zheng and Xudong Ma and Haotong Qin and Chengtao Lv and Hong Chen and Jie Luo and Xiaojuan Qi and Xianglong Liu and Michele Magno},
title = {An empirical study of LLaMA3 quantization: from LLMs to MLLMs},
year = {2024},
journal = {Visual Intelligence},
volume = {2},
url = {https://www.sciopen.com/article/10.1007/s44267-024-00070-x},
doi = {10.1007/s44267-024-00070-x},
}

@article{llama3,
  publtype={informal},
  author={Abhimanyu Dubey and Abhinav Jauhri and Abhinav Pandey and Abhishek Kadian and Ahmad Al-Dahle and Aiesha Letman and Akhil Mathur and Alan Schelten and Amy Yang and Angela Fan and Anirudh Goyal and Anthony Hartshorn and Aobo Yang and Archi Mitra and Archie Sravankumar and Artem Korenev and Arthur Hinsvark and Arun Rao and Aston Zhang and Aurélien Rodriguez and Austen Gregerson and Ava Spataru and Baptiste Rozière and Bethany Biron and Binh Tang and Bobbie Chern and Charlotte Caucheteux and Chaya Nayak and Chloe Bi and Chris Marra and Chris McConnell and Christian Keller and Christophe Touret and Chunyang Wu and Corinne Wong and Cristian Canton Ferrer and Cyrus Nikolaidis and Damien Allonsius and Daniel Song and Danielle Pintz and Danny Livshits and David Esiobu and Dhruv Choudhary and Dhruv Mahajan and Diego Garcia-Olano and Diego Perino and Dieuwke Hupkes and Egor Lakomkin and Ehab AlBadawy and Elina Lobanova and Emily Dinan and Eric Michael Smith and Filip Radenovic and Frank Zhang and Gabriel Synnaeve and Gabrielle Lee and Georgia Lewis Anderson and Graeme Nail and Grégoire Mialon and Guan Pang and Guillem Cucurell and Hailey Nguyen and Hannah Korevaar and Hu Xu and Hugo Touvron and Iliyan Zarov and Imanol Arrieta Ibarra and Isabel M. Kloumann and Ishan Misra and Ivan Evtimov and Jade Copet and Jaewon Lee and Jan Geffert and Jana Vranes and Jason Park and Jay Mahadeokar and Jeet Shah and Jelmer van der Linde and Jennifer Billock and Jenny Hong and Jenya Lee and Jeremy Fu and Jianfeng Chi and Jianyu Huang and Jiawen Liu and Jie Wang and Jiecao Yu and Joanna Bitton and Joe Spisak and Jongsoo Park and Joseph Rocca and Joshua Johnstun and Joshua Saxe and Junteng Jia and Kalyan Vasuden Alwala and Kartikeya Upasani and Kate Plawiak and Ke Li and Kenneth Heafield and Kevin Stone and et al.},
  title={The Llama 3 Herd of Models},
  year={2024},
  cdate={1704067200000},
  journal={CoRR},
  volume={abs/2407.21783},
  url={https://doi.org/10.48550/arXiv.2407.21783}
}

@article{wang-etal-2024:ACMSurvey,
author = {Wang, Song and Zhu, Yaochen and Liu, Haochen and Zheng, Zaiyi and Chen, Chen and Li, Jundong},
title = {{Knowledge Editing for Large Language Models: A Survey}},
year = {2024},
issue_date = {March 2025},
publisher = {Association for Computing Machinery},
address = {New York, NY, USA},
volume = {57},
number = {3},
issn = {0360-0300},
url = {https://doi.org/10.1145/3698590},
doi = {10.1145/3698590},
journal = {ACM Comput. Surv.},
month = nov,
articleno = {59},
numpages = {37}
}

@misc{mazzia2024surveyknowledgeeditingneural,
      title={{A Survey on Knowledge Editing of Neural Networks}}, 
      author={Vittorio Mazzia and Alessandro Pedrani and Andrea Caciolai and Kay Rottmann and Davide Bernardi},
      year={2024},
      eprint={2310.19704},
      archivePrefix={arXiv},
      primaryClass={cs.LG},
      url={https://arxiv.org/abs/2310.19704}, 
}

@article{tan2023massive,
  title={{Massive Editng for Large Language Models via Meta Learning}},
  author={Tan, Chenmien and Zhang, Ge and Fu, Jie},
  journal={arXiv preprint arXiv:2311.04661},
  year={2023}
}

@inproceedings{
guo2025balancedit,
title={{BalancEdit: Dynamically Balancing the Generality-Locality Trade-off in Multi-modal Model Editing}},
author={Dongliang Guo and Mengxuan Hu and Zihan Guan and Thomas Hartvigsen and Sheng Li},
booktitle={Forty-second International Conference on Machine Learning},
year={2025},
url={https://openreview.net/forum?id=JWtcAlXkMN}
}

@inproceedings{
wang2024wise,
title={{WISE: Rethinking the Knowledge Memory for Lifelong Model Editing of Large Language Models}},
author={Peng Wang and Zexi Li and Ningyu Zhang and Ziwen Xu and Yunzhi Yao and Yong Jiang and Pengjun Xie and Fei Huang and Huajun Chen},
booktitle={The Thirty-eighth Annual Conference on Neural Information Processing Systems},
year={2024},
url={https://openreview.net/forum?id=VJMYOfJVC2}
}

@inproceedings{youssef-etal-2025-make,
    title = "{How to Make {LLM}s Forget: On Reversing In-Context Knowledge Edits}",
    author = {Youssef, Paul  and
      Zhao, Zhixue  and
      Schl{\"o}tterer, J{\"o}rg  and
      Seifert, Christin},
    editor = "Chiruzzo, Luis  and
      Ritter, Alan  and
      Wang, Lu",
    booktitle = "Proceedings of the 2025 Conference of the Nations of the Americas Chapter of the Association for Computational Linguistics: Human Language Technologies (Volume 1: Long Papers)",
    month = apr,
    year = "2025",
    address = "Albuquerque, New Mexico",
    publisher = "Association for Computational Linguistics",
    url = "https://aclanthology.org/2025.naacl-long.630/",
    doi = "10.18653/v1/2025.naacl-long.630",
    pages = "12656--12669",
    ISBN = "979-8-89176-189-6",
}

@inproceedings{youssef-etal-2025-fact,
    title = "{Has this Fact been Edited? Detecting Knowledge Edits in Language Models}",
    author = {Youssef, Paul  and
      Zhao, Zhixue  and
      Seifert, Christin  and
      Schl{\"o}tterer, J{\"o}rg},
    editor = "Chiruzzo, Luis  and
      Ritter, Alan  and
      Wang, Lu",
    booktitle = "Proceedings of the 2025 Conference of the Nations of the Americas Chapter of the Association for Computational Linguistics: Human Language Technologies (Volume 1: Long Papers)",
    month = apr,
    year = "2025",
    address = "Albuquerque, New Mexico",
    publisher = "Association for Computational Linguistics",
    url = "https://aclanthology.org/2025.naacl-long.492/",
    doi = "10.18653/v1/2025.naacl-long.492",
    pages = "9768--9784",
    ISBN = "979-8-89176-189-6",

}

@misc{nostalgebraist2020interpreting,
  title        = {Interpreting GPT: the logit lens},
  author       = {nostalgebraist},
  year         = {2020},
  month        = {August 2020},
  howpublished = {\url{https://www.lesswrong.com/posts/AcKRB8wDpdaN6v6ru/interpreting-gpt-the-logit-lens}},
  note         = {LessWrong. Accessed: 2025-12-18}
}

@misc{cheng2025finetuningeraseeditsfragile,
      title={{Can Fine-Tuning Erase Your Edits? On the Fragile Coexistence of Knowledge Editing and Adaptation}}, 
      author={Yinjie Cheng and Paul Youssef and Christin Seifert and Jörg Schlötterer and Zhixue Zhao},
      year={2025},
      eprint={2511.05852},
      archivePrefix={arXiv},
      primaryClass={cs.CL},
      url={https://arxiv.org/abs/2511.05852}, 
}

\appendix
\section{Training Details}
\label{sec:appendix_0}
We train a single shared mask $K$ across all edits using gradient-based optimization. The mask is parameterized by learnable parameters $\Theta \in \mathbb{R}^{(m\times n)}$, with the soft mask computed as $K = \sigma(\frac \Theta \tau)$ \cite{guo_calibration}, where $\sigma$ is the sigmoid function and $\tau$ controls the sharpness of binarization.

We initialize $\Theta \sim N(0.85, 0.1)$, biasing the mask toward retaining weights initially. The temperature parameter $\tau$ starts at $6.0$ and decays with rate $3.0$ over training to encourage binary mask values. We use AdamW \cite{Loshchilov2017DecoupledWD} with learning rate $1e-3$ and $\beta = (0.9, 0.999)$ for 300 epochs. At inference, we binarize the mask using threshold $\gamma = 0.7$ for GPT-2 XL, 0.85 for Qwen-2.5, and 0.9 for LLaMa-3. The thresholds are determined as a trade-off of sparsity and the reversal success rate (RSR) for a specific model.

\paragraph{Loss hyperparameters.}For the restoration loss, we set the margin $\delta = 3.0$. For the KL divergence term, we use $\beta_{KL} = 3.26$ with temperature annealing from $T = 1.64$ to $T_{max} = 4.30$ following a linear schedule. The sparsity constraint is set to $S_{max} = 0.10$ for both model architectures and editing methods.

\paragraph{Dataset Statistics.} We use the CounterFact dataset \cite{meng2022locating} for training and evaluating the shared mask. Tables \ref{tab:rome_dataset} and \ref{tab:memit_dataset} summarize the statistics for ROME and MEMIT experiments respectively.
For ROME, we train on 3,000 single-edit samples and evaluate on 1,700 held-out samples, both stratified across 10 relation types. For MEMIT, which supports batch editing, we use 1,000 samples for training and 1,000 for testing. In the MEMIT setting, all facts within each split are edited simultaneously as a single batch edit, and the mask is trained on the last edited layer.
\begin{table}
  \centering
  \begin{tabular}{r r r}
    \toprule
    \textbf{Statistic} & \textbf{Train} & \textbf{Test}\\
    \midrule
    Facts & 3,000 & 1,700\\
    Relations & 10 & 10 \\
    Unique objects ($o_{\text{true}}$) & 236 & 204\\
    Unique subjects & 2,986 & 1,697 \\
    Unique mappings $o_{\text{true}} \to o^{*}$ & 1,284 & 866\\
    \bottomrule
  \end{tabular}
  \caption{Dataset statistics for ROME experiments using CounterFact samples.}
  \label{tab:rome_dataset}
\end{table}

\begin{table}
  \centering
  \begin{tabular}{r r r}
    \toprule
    \textbf{Statistic} & \textbf{Train} & \textbf{Test}\\
    \midrule
    Facts & 1,000 & 1,000\\
    Relations & 10 & 10 \\
    Unique objects ($o_{\text{true}}$) & 174 & 172\\
    Unique subjects & 1,000 & 1,000 \\
    Unique mappings $o_{\text{true}} \to o^{*}$ & 564 & 580\\
    \bottomrule
  \end{tabular}
  \caption{Dataset statistics for MEMIT experiments using CounterFact samples.}
  \label{tab:memit_dataset}
\end{table}
\paragraph{Computational Resources.} All experiments were conducted on an HPC cluster using NVIDIA A100 (80GB) GPUs. Mask training for ROME required approximately 40 GPU hours, while MEMIT mask training required approximately 15 GPU hours.

\section{Mask Results}
\label{sec:appendix_a}
\subsection{ROME}
\begin{figure*}
    \centering
    \includegraphics[width=\textwidth,clip, trim={5.2cm 0.2cm 6.9cm 0.2cm}]{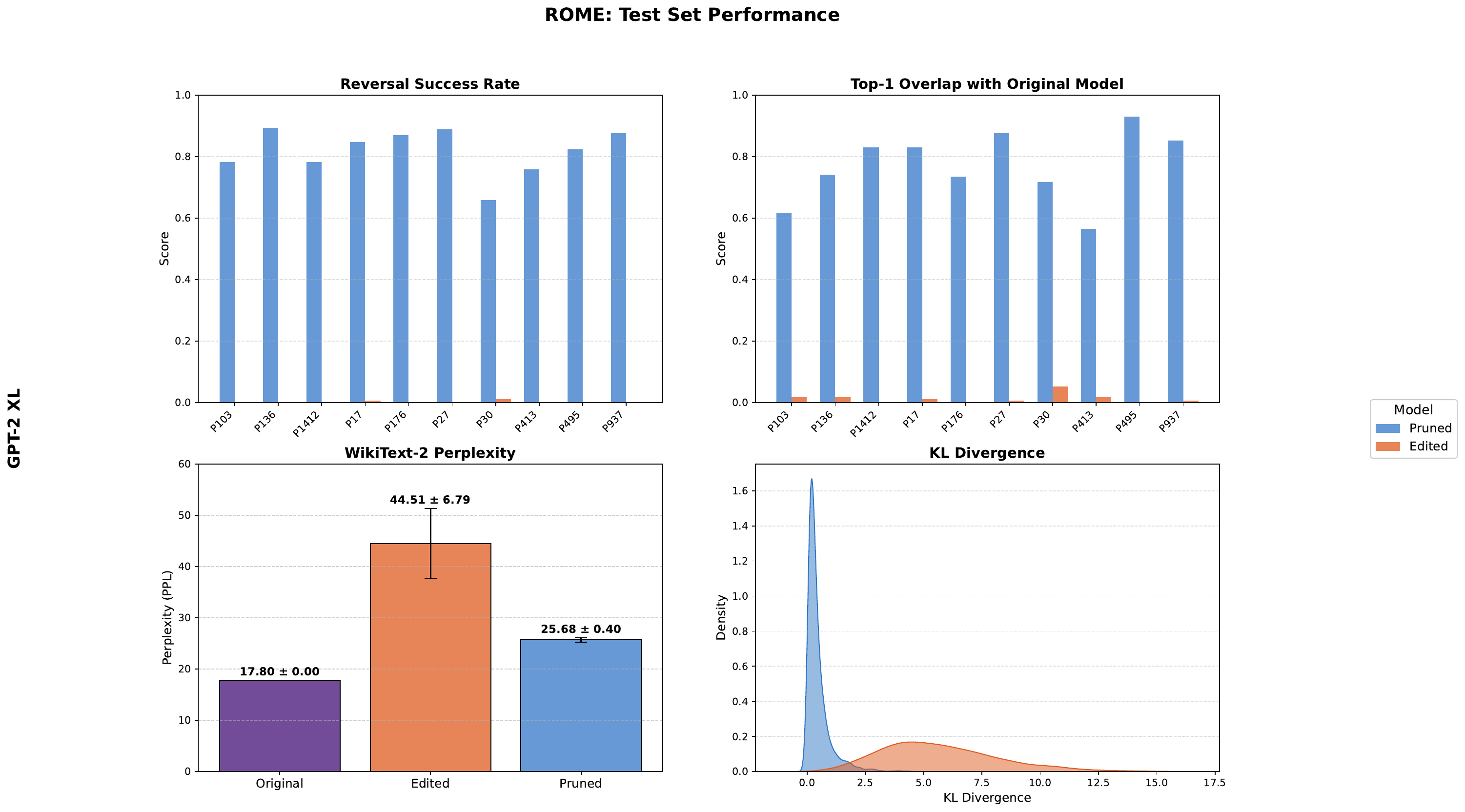}
    \caption{
        \textbf{ROME's Reversal performance in GPT-2 XL.}
        First Row: We report the Reversal Success Rate (left) and Top-1 Overlap with the original model (right) across different relation types (e.g., P103, P136).
        Second Row: The leftmost plot compares WikiText-2 Perplexity, demonstrating that the pruned model ($M_p$) significantly reduces the perplexity degradation caused by the edit ($M_e$), recovering capabilities closer to the original model ($M$). On the right, we report the KL-divergence between 2 pairs of model states: 1) the original $M$ and the edited $M_e$ (orange); 2) the original $M$ and the pruned $M_p$ (blue), showing that pruned model is closer to the original model than the edited one.
    }
    \label{fig:rsr_results}
\end{figure*}
\autoref{fig:rsr_results} presents detailed reversal performance for ROME edits on GPT-2 XL. The Reversal Success Rate (RSR) remains consistently high across all relation types, ranging from approximately 75\% to 90\%, demonstrating that the learned mask generalizes across semantically diverse facts. Top-1 Overlap follows a similar pattern, confirming that the pruned model not only prefers the original fact but also recovers the exact prediction behavior of the unedited model in most cases.

The perplexity analysis reveals that ROME edits substantially degrade language modeling capabilities, with mean perplexity increasing from 17.80 to 44.51. Applying the shared mask reduces perplexity to 25.68, recovering much of the lost performance. The KL-divergence distribution further confirms that the pruned model's output distribution is substantially closer to the original model than the edited model, with the majority of samples showing lower divergence after mask application.

In more extreme cases, ROME edits can cause model collapse \cite{yang-etal-2024-butterfly}, leading to perplexity spikes ranging from hundreds to tens of millions. As illustrated in \autoref{tab:repair_examples}, our learned mask is able to substantially recover performance in these scenarios, reducing perplexity by several orders of magnitude without modifying the edited weights themselves.
\begin{table}
    \centering
    \small
    \begin{tabular}{r r r}
    \toprule
    \textbf{Case ID} & \textbf{PPL ($M_e$)} & \textbf{PPL ($M_p$)} \\
    \midrule
    3877  & 21,840,210.0 & 55.0 \\
    13259 & 1,195.8      & 32.0 \\
    16110 & 1,091.5      & 31.5 \\
    102   & 969.1        & 34.3 \\
    20421 & 748.3        & 35.0 \\
    \bottomrule
    \end{tabular}
    \caption{
        \textbf{Examples of perplexity recovery.} 
        Selected cases where the initial edit ($M_e$) caused catastrophic perplexity spikes, which were significantly repaired by the shared mask ($M_p$).
    }
    \label{tab:repair_examples}
\end{table}

\subsection{MEMIT}
\begin{figure*}
    \centering
    \includegraphics[width=\textwidth]{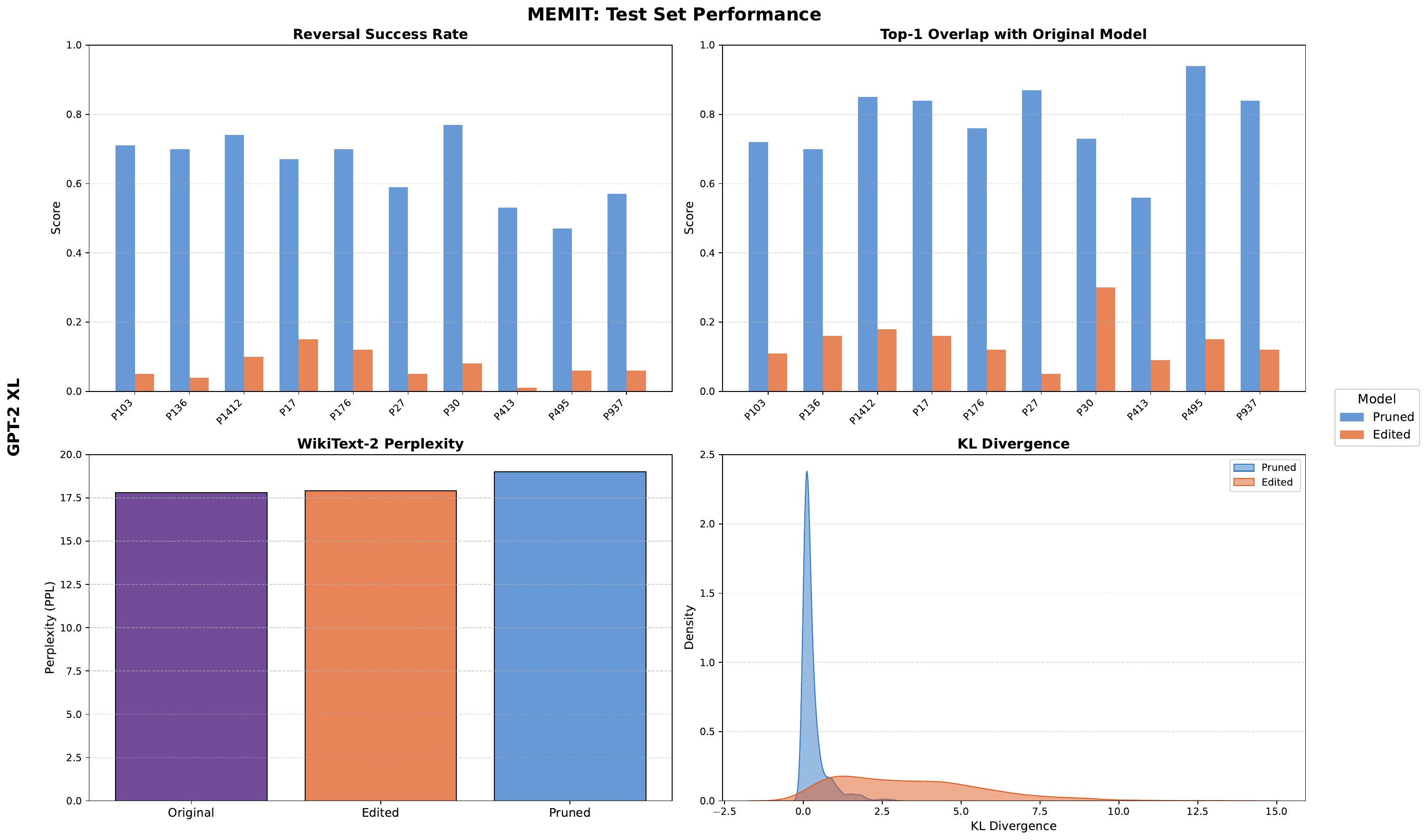}
    \caption{
        \textbf{MEMIT's Reversal performance in GPT-2 XL.}
        First Row: Reversal Success Rate (left) and Top-1 Overlap (right) across relation types, showing consistent reversal performance despite MEMIT's multi-layer editing.
        Second Row: WikiText-2 Perplexity (left) remains close to the original after both editing and pruning, indicating MEMIT causes less collateral damage than ROME. KL-divergence (right) confirms the pruned model distribution is closer to the original than the edited model.
    }
    \label{fig:memit_results}
\end{figure*}
\autoref{fig:memit_results} presents the corresponding analysis for MEMIT edits. Despite MEMIT distributing edits across multiple connected layers, the shared mask trained only on the last edited layer achieves comparable reversal performance to ROME. RSR remains above 70\% for most relation types, with Top-1 Overlap showing similar consistency.

A notable difference from ROME is that MEMIT edits cause minimal perplexity degradation: WikiText-2 perplexity increases only marginally from 17.80 to 17.90 on GPT-2 XL. This suggests that MEMIT's distributed editing strategy is less disruptive to general language modeling capabilities. However, the KL-divergence analysis reveals that despite this stability, MEMIT edits still shift the model's output distribution away from the original, and the mask successfully reduces this divergence.

\subsection{Generalization to Unseen Relations}
To assess whether the mask generalizes beyond the relation set used in training and test sets, we evaluate it on 1,000 ROME edits spanning 10 additional CounterFact relations that were not seen during mask training. The mask is applied without any re-training or fine-tuning. If the mask had memorized relation-specific structure, we would expect a substantial drop in reversal performance. Instead, as \autoref{tab:ood} shows, the mask maintains strong reversal on OOD relations, achieving 77\% RSR on GPT-2 XL and 59\% on LLaMA-3.2. This supports our central claim that the targeted subspace reflects a common mechanism shared across semantically diverse facts rather than a relation-specific pattern.

\begin{table}[h]
\centering
\small
\begin{tabular}{lc}
\toprule
Model & OOD RSR $\uparrow$ \\
\midrule
GPT-2 XL        & 77\% \\
LLaMA-3.2 (3B)  & 59\% \\
\bottomrule
\end{tabular}
\caption{\textbf{OOD generalization of the ROME mask.}
Reversal Success Rate on 1{,}000 ROME edits spanning 10 CounterFact OOD relations. The mask is applied without any re-training. Strong RSR on unseen relations indicates that the identified subspace is not tied to the relations used during mask training.}
\label{tab:ood}
\end{table}

\section{MEMIT Analysis: Decomposition of Residual Stream}
\label{sec:appendix_b}

We extend the residual stream analysis from Section~5 to MEMIT edits. While ROME modifies a single layer, MEMIT distributes edits across multiple consecutive layers. This raises the question: does MEMIT exploit the same overattention mechanism as ROME, or does its distributed editing strategy produce fundamentally different internal dynamics?

\subsection{Edits Induce Overattention}

\begin{figure*}
    \centering
    \begin{subfigure}{0.48\linewidth}
        \centering
        LLaMA-3.2 (3B) \\ \vspace{0.2em}
        \includegraphics[width=\linewidth]{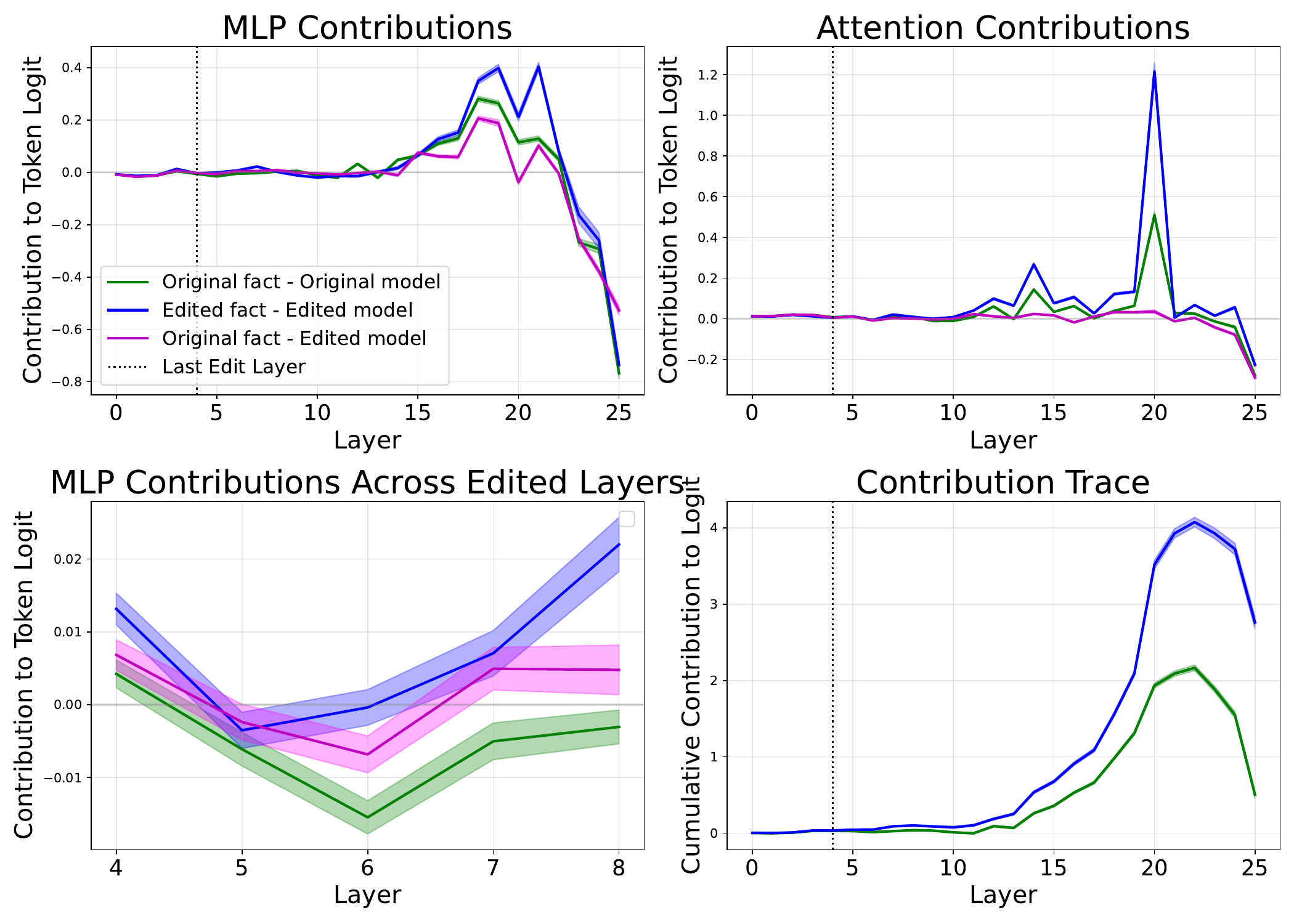}
    \end{subfigure}
    \hfill
    \begin{subfigure}{0.48\linewidth}
        \centering
        GPT-2 XL \\ \vspace{0.2em}
        \includegraphics[width=\linewidth]{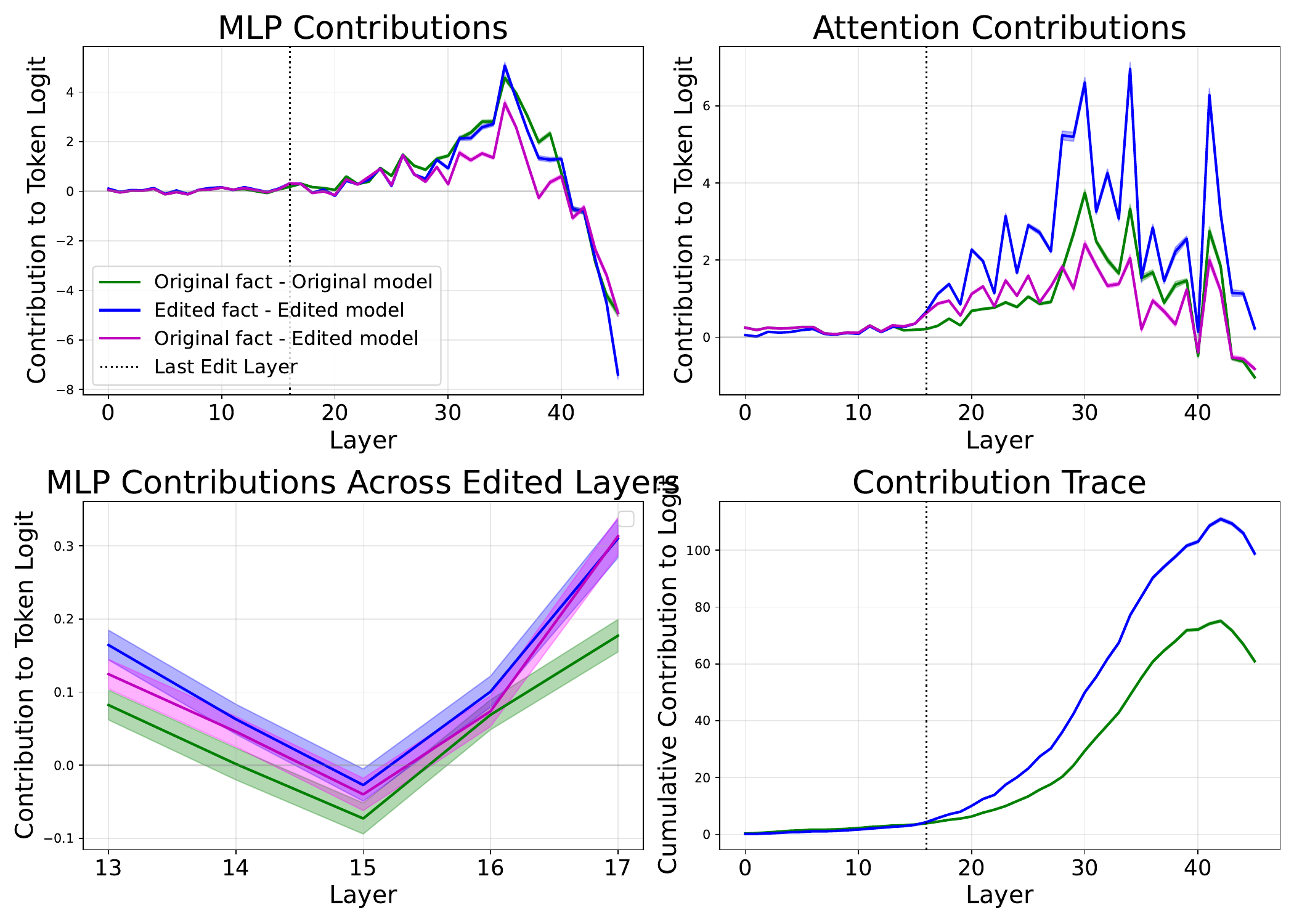}
    \end{subfigure}
    \caption{\textbf{Residual stream decomposition for MEMIT edits.} 
        Top row: MLP (left) and attention (right) contributions per layer. 
        Bottom row: MLP contributions across the edited layers only (left) and cumulative logit contribution trace (right).
        Green: original fact in original model; blue: edited fact in edited model; purple: original fact in edited model. 
        Both architectures show amplified attention contributions in downstream layers, consistent with the overattention mechanism identified for ROME.
    }
    \label{fig:memit_decomposition_edited}
\end{figure*}

\autoref{fig:memit_decomposition_edited} presents the residual stream decomposition for MEMIT edits. Despite the distributed nature of MEMIT, we observe a similar overattention pattern identified for ROME.

\paragraph{GPT-2 XL.} The MLP contributions across the edited layers (13--17) show modest amplification compared to the original model, with the edited model giving higher contributions for edited (blue) and original (purple) facts than the original fact in the unedited model (green). However, the dominant effect emerges in the attention contributions: a sharp spikes appear in the downstream layers, substantially exceeding the original model's attention pattern. The cumulative contribution trace confirms this: edited facts accumulate dramatically higher logit contributions, with the gap widening primarily through downstream attention layers.

\paragraph{LLaMA-3.2 (3B).} The pattern differs in timing but not in mechanism. MLP contributions across the edited layers (4--8) show minimal differentiation between conditions. Instead, attention contributions exhibit a pronounced spike in later layers (17--22), consistent with ROME's delayed overattention effect on this architecture. The cumulative trace shows edited facts reaching substantially higher final contributions.

\paragraph{Key difference from ROME.} Unlike ROME, which injects a large signal at a single layer (35$\times$ amplification in GPT-2 XL), MEMIT's distributed edits produce smaller per-layer perturbations. However, these perturbations compound through downstream attention, ultimately producing comparable overattention effects. 

\subsection{Mask Eliminates Overattention}

\begin{figure*}
    \centering
    \begin{subfigure}{0.48\linewidth}
        \centering
        LLaMA-3.2 (3B) \\ \vspace{0.2em}
        \includegraphics[width=\linewidth]{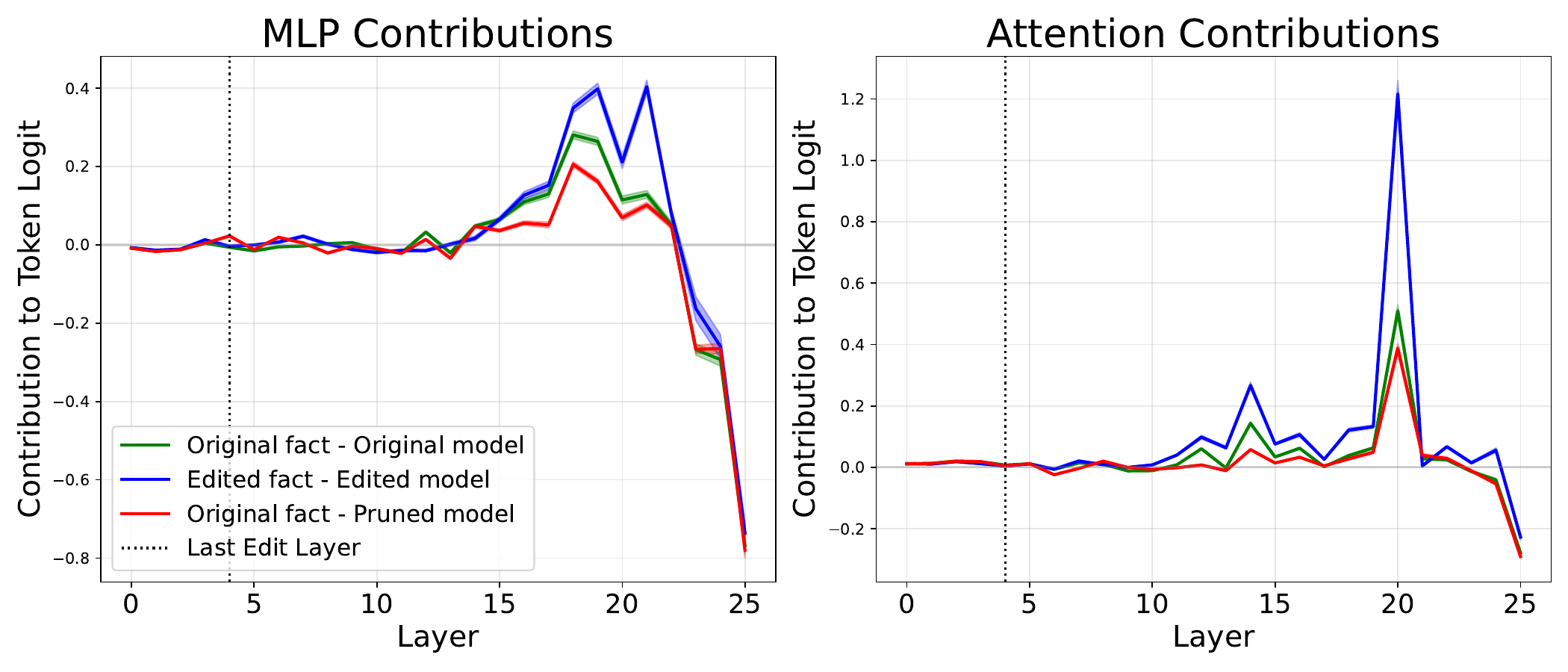}
    \end{subfigure}
    \hfill
    \begin{subfigure}{0.48\linewidth}
        \centering
        GPT-2 XL \\ \vspace{0.2em}
        \includegraphics[width=\linewidth]{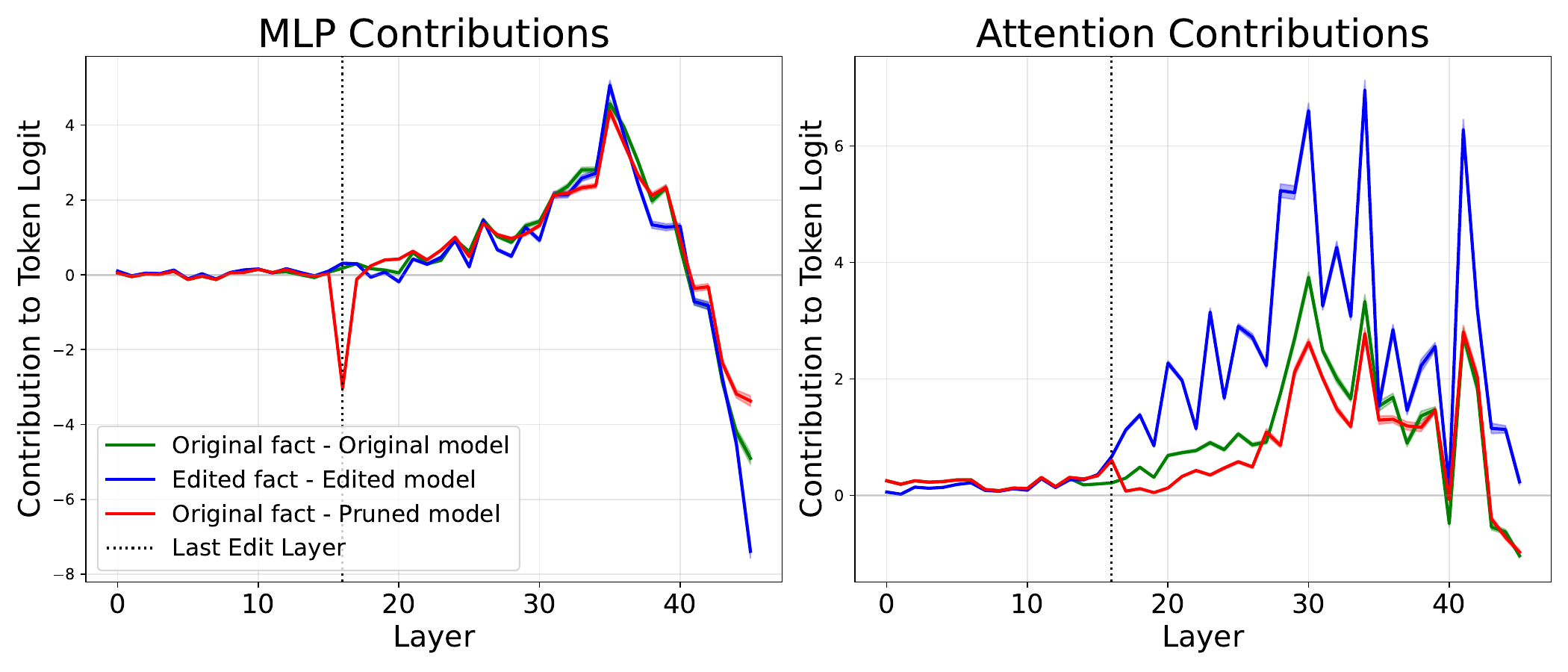}
    \end{subfigure}
    \caption{\textbf{Effect of the learned mask on MEMIT edits.}
        MLP contributions (left) and attention contributions (right) for the original model (green), edited model (blue), and pruned model (red).
        The mask eliminates the amplified attention contributions visible in \autoref{fig:memit_decomposition_edited} while preserving MLP trajectories close to the original model.
    }
    \label{fig:memit_decomposition_pruned}
\end{figure*}

\autoref{fig:memit_decomposition_pruned} shows the effect of applying the learned mask to MEMIT-edited models. The mask, trained only on a single edited layer, successfully eliminates the overattention pattern while preserving MLP dynamics.

\paragraph{GPT-2 XL.} The attention spike in layers 25--35 is substantially reduced in the pruned model (red), returning toward the original model's trajectory (green). Critically, MLP contributions remain largely unaffected by the mask, following similar paths for both the original and pruned models. The key difference is that mask significantly reduces the contribution in the edited MLP block, counterbalancing the effects of editing. 

\paragraph{LLaMA-3.2 (3B).} The late-layer attention spike (layers 17--22) is eliminated in the pruned model. The MLP contributions show close alignment between the original and pruned models throughout all layers, including those beyond the edit site.

\paragraph{Consistency with ROME.} The mask's effect on MEMIT mirrors its effect on ROME: eliminating downstream overattention while preserving MLP pathways. This convergence provides strong evidence that \textbf{ROME and MEMIT exploit the same shared mechanism}, and that this mechanism can be targeted by a single sparse mask regardless of whether edits are concentrated in one layer or distributed across several.

\section{Mask analysis and Pruning}
\label{sec:appendix_c}

\subsection{Mask Analysis}
\begin{figure}[H]
    \centering
    \includegraphics[width=0.7\linewidth]{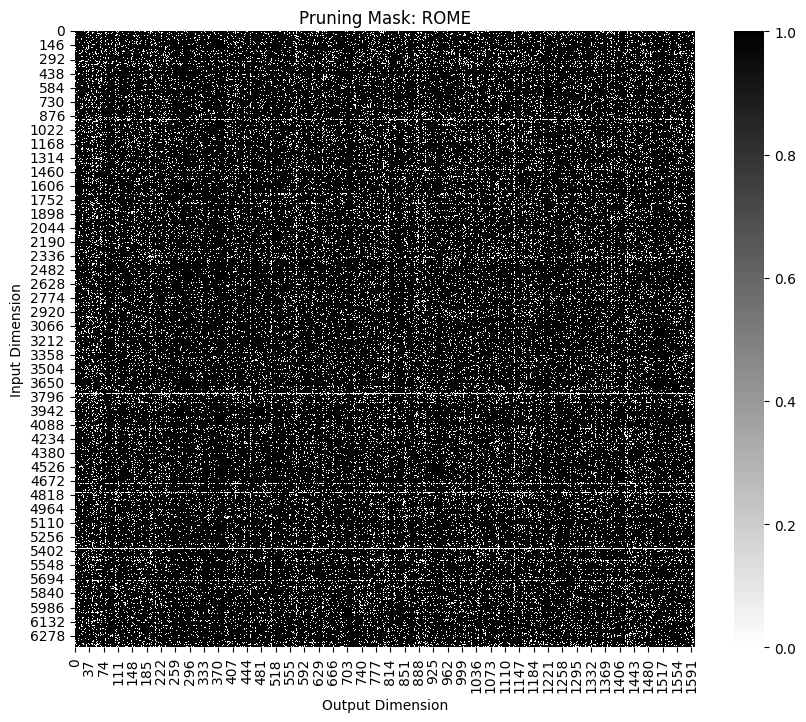}
    \caption{\textbf{Learned mask structure.} Heatmap of the binary mask over the edited MLP weight matrix in GPT-2 XL. White denotes pruned weights (mask value 0), black denotes retained weights (mask value 1). The mask exhibits column-wise sparsity to a limited extent, targeting specific output dimensions.}
    \label{fig:mask_heatmap}
\end{figure}

To understand how the mask reverses edits, we analyze which components it targets. \autoref{fig:mask_heatmap} visualizes the learned mask as a heatmap over the edited weight matrix. The mask does not uniformly prune weights or remove entire neurons; instead, it exhibits a structured sparsity pattern, concentrating on specific output dimensions (columns) of the MLP weight matrix while leaving others largely intact.

\begin{figure}[H]
    \centering
    \includegraphics[width=0.7\linewidth]{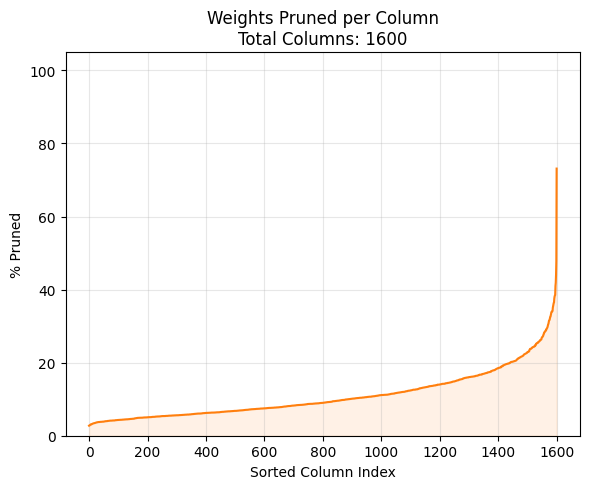}
    \caption{\textbf{Distribution of pruned weights per output dimension.} Each point represents one of the 1,600 output dimensions (columns) of the edited MLP weight matrix, sorted by pruning percentage. The steep rise on the right indicates that pruning is concentrated in a small subset of dimensions, with most dimensions retaining over 90\% of their weights.}
    \label{fig:weights_per_column}
\end{figure}

\autoref{fig:weights_per_column} quantifies this column-wise concentration. The distribution is highly skewed: the majority of output dimensions have only a small subset of weights pruned, while some dimensions are heavily targeted. The top-5 most pruned dimensions have over 40\% of their weights pruned, with dimension 214 reaching 74.6\%. This concentrated pruning pattern suggests the mask identifies specific functional pathways critical for maintaining edits.
Notably, only 18\% of pruned weights correspond to high-magnitude ROME updates, suggesting the mask targets functional pathways rather than simply reversing the largest weight changes. \autoref{tab:delta_magnitude} confirms this: defining the update magnitude as $\Delta \hat{W} = \hat{W} - W$, i.e., the difference between edited and original weights, the mean $\|\Delta \hat{W}\|$ at masked positions is an order of magnitude smaller than the overall mean $\|\Delta \hat{W}\|$, indicating that the mask prunes structurally important but not exclusively high-magnitude weight updates.

\begin{table}[h]
\centering
\small
\begin{tabular}{lcc}
\toprule
\textbf{Model} & \textbf{Edited $\|\Delta \hat{W}\|$} & \textbf{Masked $\|\Delta \hat{W}\|$} \\
\midrule
GPT-2 XL      & 18.47 $\pm$ 0.040 & 1.85 $\pm$ 0.005 \\
LLaMA-3.2 3B  & 3.91 $\pm$ 0.009  & 0.57 $\pm$ 0.002 \\
\bottomrule
\end{tabular}
\caption{Mean $\|\Delta \hat{W}\|$ magnitude at all edited positions versus positions 
targeted by the mask.}
\label{tab:delta_magnitude}
\end{table}

\begin{figure*}[t]
    \centering
    {GPT-2 XL} \\
    \includegraphics[width=\linewidth]{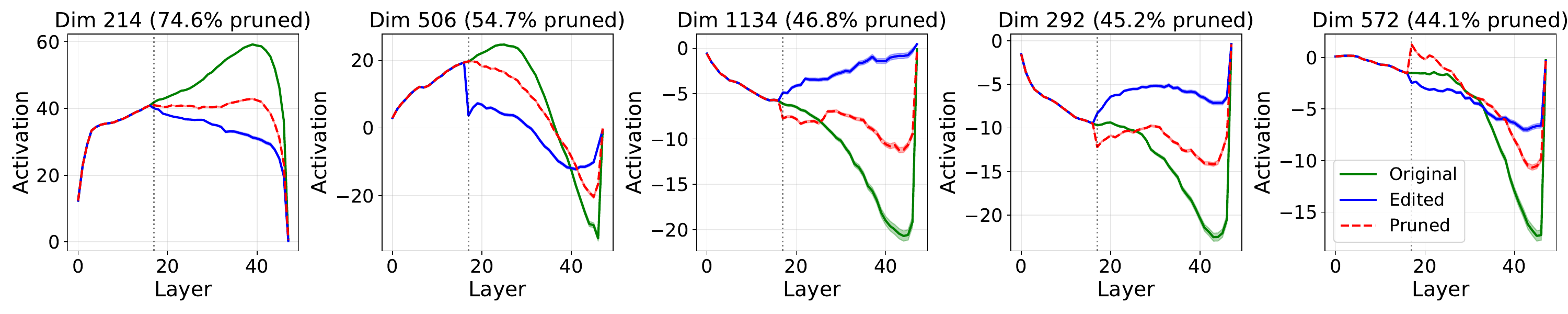}
    
    \vspace{0.2cm}
    
    {Llama-3 3B} \\
    \includegraphics[width=\linewidth]{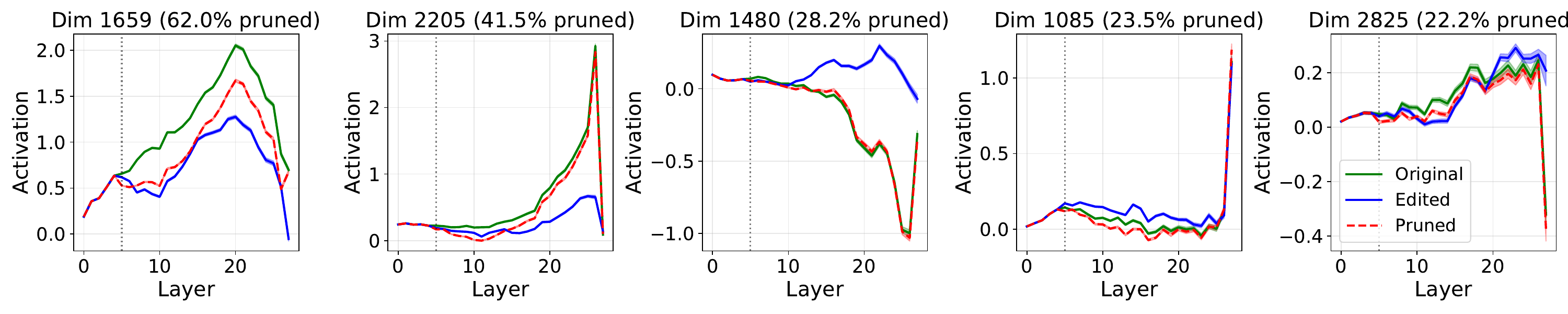}
    
    \caption{\textbf{Activation trajectories of the top-5 most pruned dimensions.} 
    Mean activation (with standard error) across 1,000 samples tracked through all layers of GPT-2 XL and Llama-3 3B. 
    Green: original model; blue: edited model; red dashed: pruned model.}
    \label{fig:top-5_pruned}
\end{figure*}

\paragraph{Tracking pruned dimensions across layers.} To verify that the heavily pruned dimensions are indeed functionally relevant, we track their activation trajectories across layers in the original, edited, and pruned models (\autoref{fig:top-5_pruned}). 

Across all five dimensions, the edited model (blue) diverges substantially from the original GPT2-XL trajectory (green) after the edit layer. Dimension 214, the most heavily pruned (74.6\%), shows the clearest effect: its activation is suppressed by the edited model. The remaining dimensions (506, 1134, 292, 572) each show distinct divergence patterns where the edited model deviates from the original baseline.

The same pattern holds for LLaMA-3.2. Dimension 1659, the most heavily pruned (62.0\%), shows clear divergence after the edit layer, with the edited model suppressing its activation relative to the original. The remaining dimensions (2205, 1480, 1085, 2825) similarly exhibit distinct divergence patterns, confirming that the mask targets functionally relevant pathways across architectures.

In all cases, applying the mask (red dashed) restores the activation trajectories toward the original model. This consistent restoration across the most heavily pruned dimensions in both GPT-2 XL and LLaMA-3 confirms that the mask precisely targets the activation changes induced by editing, reversing their effect on downstream computation rather than introducing arbitrary perturbations.

\subsection{Comparison with Traditional Pruning Methods}
\label{sec:appendix_pruning}

To contextualize our learned mask approach, we compare against traditional pruning methods. We evaluate four pruning criteria: (1) unstructured magnitude pruning on the weight difference $\Delta W = W - \hat{W}$, (2) unstructured magnitude pruning on the edited weights $\hat{W}$ directly, (3) structured magnitude pruning based on column norms, and (4) structured activation-based pruning using average $W_{fc}$ activations. Each criterion is tested in two modes: \textit{zero} (pruned weights set to 0) and \textit{original} (pruned weights restored to pre-edit values).

\begin{figure*}
    \centering
    \includegraphics[width=1.0\linewidth]{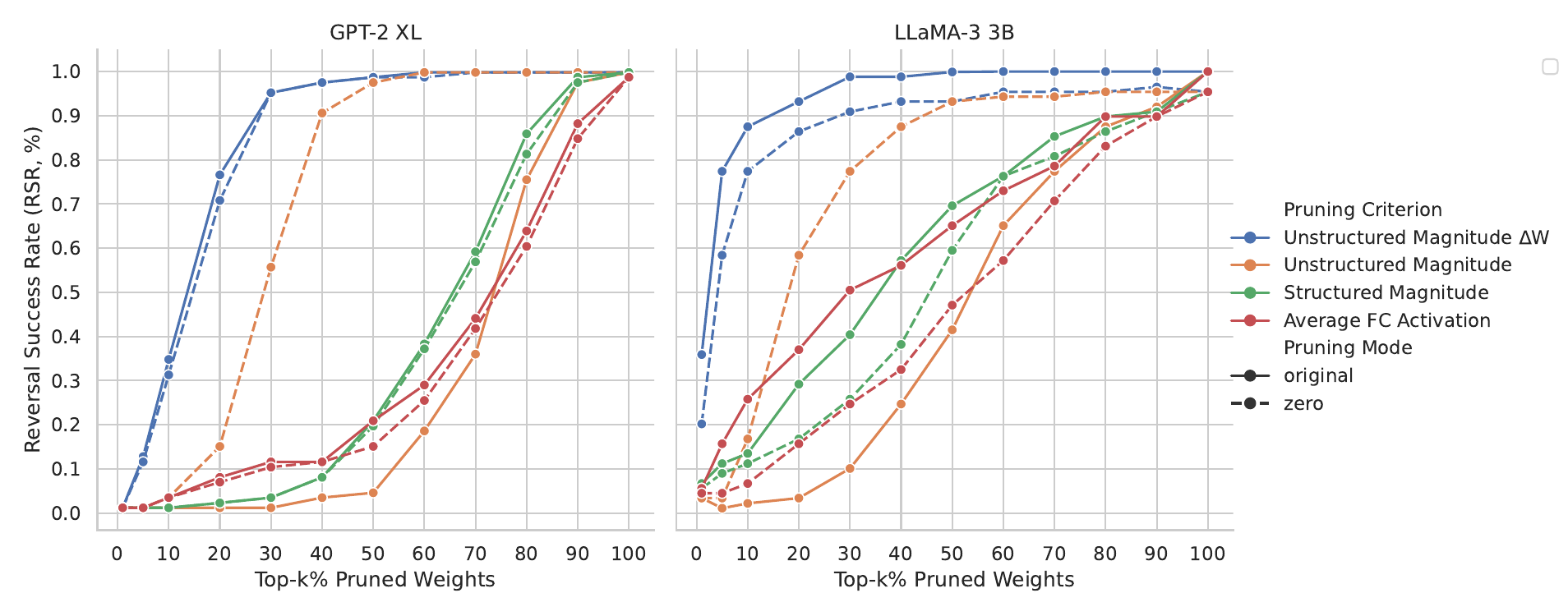}
    \caption{\textbf{Comparison of traditional pruning methods.} Reversal Success Rate (RSR) as a function of pruning percentage for GPT-2 XL (left) and LLaMA-3 3B (right). Each curve corresponds to a different pruning criterion. Solid lines indicate the \textit{original} mode (pruned weights restored to pre-edit values); dashed lines indicate the \textit{zero} mode (pruned weights set to 0). Unstructured $\Delta W$ pruning achieves 90\% RSR with only 30\% pruning, while structured methods require near-complete removal of the edited layer.}
    \label{fig:pruning}
\end{figure*}

\autoref{fig:pruning} shows the Reversal Success Rate (RSR) as a function of pruning percentage. The results reveal a clear hierarchy among methods. Unstructured $\Delta W$ pruning is most efficient: zeroing only 30\% of the weights with the largest update magnitude achieves over 90\% RSR on both architectures. Unstructured magnitude pruning on $\hat{W}$ requires substantially more intervention (approximately 40\% for GPT-2 XL and 50\% for LLaMA-3 (3B)) and shows a notable asymmetry between modes: zero mode is far more effective than restoring original values.

Structured pruning methods perform considerably worse. Both structured magnitude and activation-based criteria show a near-linear relationship between pruning rate and RSR, requiring 90--100\% of weights to be removed before reaching 90\% reversal. This indicates that the edit's effect cannot be attributed to a small number of neurons; rather, it is distributed across the weight matrix in a way that structured approaches cannot efficiently target.

These findings motivate our mask training approach: while $\Delta W$ pruning provides a strong baseline, it still requires removing 30\% of weights. Our learned masks achieve comparable or better reversal performance while pruning less than 10\% of weights (\autoref{tab:reversal_detailed}), demonstrating that optimization can identify more precise targets than magnitude-based heuristics.

\end{document}